\documentclass[11pt]{article}

\usepackage[final]{acl}

\usepackage{times}
\usepackage{latexsym}

\usepackage[T1]{fontenc}
\usepackage[utf8]{inputenc}

\usepackage{microtype}

\usepackage{inconsolata}

\usepackage{graphicx}
\usepackage{booktabs}
\usepackage{makecell}
\usepackage{multirow}
\usepackage{colortbl}
\usepackage{threeparttable}
\usepackage{tabularx}
\usepackage{pifont}
\usepackage{subcaption}
\usepackage[most]{tcolorbox}
\usepackage{xcolor}    % 用于颜色定义
\usepackage{calc}
\usepackage{amsmath} % 提供更好的数学支持
\usepackage{array}
\usepackage{pgf}
\tcbuselibrary{skins, breakable}
\usepackage{longtable}
\newcommand{\llm}[1]{\textsc{#1}}
\newcommand{\model}[1]{\texttt{#1}}

\title{Persona-E$^2$: A Human-Grounded Dataset for Personality-Shaped Emotional Responses to Textual Events}

\author{
  Yuqin Yang \quad Haowu Zhou \quad Haoran Tu \quad Zhiwen Hui \quad Shiqi Yan \\
  \textbf{Haoyang Li \quad Dong She \quad Xianrong Yao \quad Yang Gao \quad Zhanpeng Jin$^{\dagger}$} \\
  School of Future Technology \\
  South China University of Technology, Guangzhou, China \\
  \texttt{\{ftyuqin\_yang, 202364870491, 202330691461, 202364870731\}@mail.scut.edu.cn} \\
  \texttt{\{202364871202, ftlhy, ftdshe, ftxryao\}@mail.scut.edu.cn} \\
  \texttt{\{gaoyang2025, zjin\}@scut.edu.cn} \\
}

\tcbset{
    promptstyle/.style={
        colframe=gray!50!black, % 深灰色边框
        rounded corners,        % 圆角
        left=8pt, right=8pt, top=8pt, bottom=8pt, % 适当的内边距，增加可读性
        fonttitle=\bfseries,    % 标题加粗
        fontupper=\footnotesize,% 自动将内容设置为 footnotesize，无需手动在正文里写
        breakable,              % 允许跨页（对于长列表非常重要）
        enhanced,               % 启用高级绘图引擎（优化圆角和跨页效果）
        coltitle=white,         % 标题文字颜色
        attach boxed title to top left={xshift=10pt, yshift*=-\tcboxedtitleheight/2}, 
        boxed title style={colback=gray!50!black} % 如果你想用“挂在边框上”的标题风格，取消注释这两行，否则使用默认标题栏
    }
}

\begin{document}
\maketitle
\begin{abstract}

Most affective computing research treats emotion as a static property of text, focusing on the writer's sentiment while overlooking the reader's perspective. This approach ignores how individual personalities lead to diverse emotional appraisals of the same event. Although role-playing Large Language Models (LLMs) attempt to simulate such nuanced reactions, they often suffer from ``personality illusion''---relying on surface-level stereotypes rather than authentic cognitive logic. A critical bottleneck is the absence of ground-truth human data to link personality traits to emotional shifts. To bridge the gap, we introduce \textbf{Persona-E\textsuperscript{2}} (Persona-Event2Emotion), a large-scale dataset grounded in annotated MBTI and Big Five traits to capture reader-based emotional variations across news, social media, and life narratives. Extensive experiments reveal that state-of-the-art LLMs struggle to capture precise appraisal shifts, particularly in social media domains. Crucially, we find that personality information significantly improves comprehension, with the Big Five traits alleviating ``personality illusion.''

\end{abstract}

\section{Introduction}

\begin{figure*}[t]
    \centering
    \includegraphics[width=\textwidth]{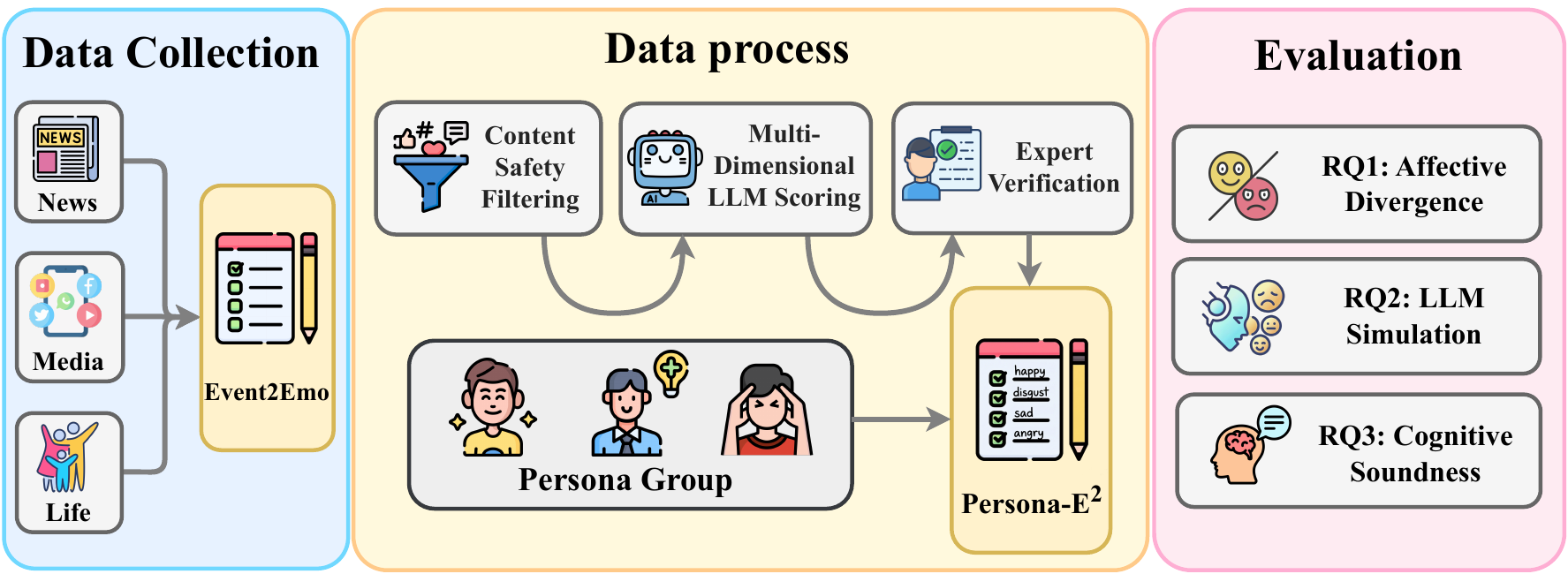} 
 \caption{Overview of the Persona-E\textsuperscript{2} framework. Events from three domains undergo multi-stage data processing. High-quality stimuli are then annotated by a Persona Group, serving to evaluate three research questions.}
    \label{fig:flow_chart}
\end{figure*}

\textit{``Two individuals can construe their situations quite similarly (agree on all the facts), and yet react with very different emotions, because they have appraised the adaptational significance of those facts differently.''}~\cite{1991Emotion}

The study of affective appraisal of events has long been central to affective computing and cognitive psychology~\cite{plaza2024emotion}. While appraisal theories suggest that emotions emerge through individualized appraisals shaped by goals and dispositions~\cite{1991Emotion,scherer1994evidence}, NLP research has largely focused on writer-expressed sentiments and reader-based unified emotional labels~\cite{plaza2024emotion}. This focus overlooks reader-based nuanced perception~\cite{buechel2017readers}, which is critical for applications, including empathetic agents, mental health support, and personalized AI assistants, that must not only process the texts but also reason about how different individuals appraise the same event diversely.

Recent interest in role-playing LLMs aims to simulate individualized reactions by injecting rich personality profiles into prompts~\cite{tseng2024two, chen2024oscars, hu2024quantifying, mao2024editing}. Despite this promise, these methods often exhibit ``personality illusion''~\cite{han2025personality}: models tend to imitate stereotypical behaviors rather than adopting the cognitive appraisal patterns based on personality. Crucially, LLM-generated labels lack grounding in authentic feedback~\cite{li2025big5}, making them insufficient for evaluating whether models truly capture emotional diversity~\cite{samuel2024personagym}. Thus, the field still lacks a human-grounded dataset to validate and enhance personality-conditioned emotion elicitation.

To address the gap, we introduce a novel dataset, \textbf{Persona-E\textsuperscript{2}} (Persona-Event2Emotion), which incorporates the popular Myers-Briggs Type Indicator (MBTI)~\cite{myers1962myers, john1991big} and the robust Big Five Inventory (BFI)~\cite{john2010handbook} traits into reader-based emotion labeling. As shown in Fig.~\ref{fig:flow_chart} by engaging annotators with assessed personality profiles to label events across diverse domains (News, Social Media, Life Experience narratives), Persona-E\textsuperscript{2} enables a controlled analysis of the personality effect on the appraisals of identical textual events~\cite{troiano2023dimensional}. Notably, unlike previous corpora, Persona-E\textsuperscript{2} prioritizes annotation density (36 labels per event) to capture diverse, trait-shaped responses (Tab.~\ref{tab:DatasetCompare}).

To evaluate the utility of Persona-E\textsuperscript{2}, we address three key research questions, through the experimental design in Sec.~\ref{sec:Experiment}:
\begin{itemize}
\item \textbf{RQ1. Affective Divergence:} How do emotional responses diverge across the General Writer, General Reader, and Persona Reader, and how is this variance modulated by source domain and personality traits?
\item \textbf{RQ2. LLM Simulation:} Can LLMs effectively simulate Persona Reader responses, particularly when faced with elicitation conflicts? 
\item \textbf{RQ3. Cognitive Soundness:} Do LLMs generate psychologically grounded rationales for their predictions, and what methods can enhance their cognitive validity? 
\end{itemize}

Our analysis reveals that affective appraisal is a domain-sensitive process, with disagreement serving as a structured personality signal. While LLMs struggle to predict precise appraisal shifts, particularly in social media domains, personality traits improve LLMs' comprehension, and BFI outperforms MBTI in mitigating ``personality illusion.'' We publicly release the dataset to foster further research in the community. Further details are provided on our project page  \url{https://scut-hai.github.io/Persona-E2/}, with the data available for download\footnote{\url{https://huggingface.co/datasets/CRIS-Yang/Persona-E2-Dataset}}\footnote{\url{https://www.kaggle.com/datasets/crisyang777/peronsa-e-personality-shaped-emotion-dataset}}.

\definecolor{row-green}{RGB}{235,250,238}
\definecolor{row-yellow}{RGB}{255,249,221}
\definecolor{row-pink}{RGB}{254,235,242}
\definecolor{row-blue}{RGB}{231,242,255}
\definecolor{kellygreen}{rgb}{0.3, 0.73, 0.09}
\definecolor{alizarin}{rgb}{0.82, 0.1, 0.26}
\definecolor{royalblue}{rgb}{0.25,0.41,1}
\newcommand{\cmark}{{\color{kellygreen} \ding{51}}}
\newcommand{\xmark}{{\color{alizarin} \ding{55}}}

\begin{table*}[!htbp]
  \centering
  \resizebox{\textwidth}{!}{
  \begin{tabular}{lcccccc}
    \toprule
    \textbf{Dataset} & \textbf{Year} & \textbf{\#Events}  & 
    \textbf{\#Annotations} & \textbf{Perspective} & \textbf{\#Emotions} & \textbf{Personality}  \\
    \midrule

    %%%%%%%%%%%%%%%%%%%%%%%%%%%%%%%
    % NEWS SECTION
    %%%%%%%%%%%%%%%%%%%%%%%%%%%%%%%
    \rowcolor{row-green}
    \multicolumn{7}{l}{{\textit{\textbf{News-based Domain}}}} \\

    \rowcolor{row-green}
    GoodNewsEveryone~\cite{oberlander2020goodnewseveryone} & 2020 & 5,000 & 15,000 & Writer + Reader & 15 & \xmark\\

    \rowcolor{row-green}
    NewsMTSC~\cite{hamborg2021newsmtsc} & 2021 & 11,029 & 56,000 & Writer & 7 & \xmark\\

    \rowcolor{row-green}
    iNews~\cite{hu2025inews} & 2025 & 2,899 & 14,550 & Reader & 6 & \xmark\\

    %%%%%%%%%%%%%%%%%%%%%%%%%%%%%%%
    % SOCIAL MEDIA SECTION
    %%%%%%%%%%%%%%%%%%%%%%%%%%%%%%%
    \rowcolor{row-yellow}
    \multicolumn{7}{l}{{\textit{\textbf{Social Media Domain}}}} \\
    
    \rowcolor{row-yellow}
    SemEval-2018 Task 1~\cite{mohammad-etal-2018-semeval} & 2018 & 22,000 & 700,000 & Writer & 4 & \xmark\\

    \rowcolor{row-yellow}
    GoEmotions~\cite{demszky2020goemotions} & 2020 & 58,000 & 118,000 & Reader & 27 & \xmark\\

    \rowcolor{row-yellow}
    SMP2020-EWECT~\cite{bertsmp2020ewect} & 2020 & 34,768 & 36,374 & Writer & 6 & \xmark\\

    \rowcolor{row-yellow}
    SenWave~\cite{yang2025senwave} & 2025 & 10,000 & 20,000 & Writer & 10 & \xmark\\

    %%%%%%%%%%%%%%%%%%%%%%%%%%%%%%%
    % LIFE EXPERIENCE SECTION
    %%%%%%%%%%%%%%%%%%%%%%%%%%%%%%%
    \rowcolor{row-pink}
    \multicolumn{7}{l}{{\textit{\textbf{Life Experience Domain}}}} \\

    \rowcolor{row-pink}
    ISEAR~\cite{scherer1994evidence} & 1994 & 7,666 & 7,666 & Writer & 7 & \xmark \\

    \rowcolor{row-pink}
    Event2Mind~\cite{rashkin2018event2mind} & 2018 & 24,716 & 57,000 & Writer + Reader & OV & \xmark \\

    \rowcolor{row-pink}
    ATOMIC~\cite{sap2019atomic} & 2019 & 24,000 & 72,000 & Experiencer & OV & \xmark \\

    \rowcolor{row-pink}
    EmpatheticDialogues~\cite{rashkin2019towards} & 2019 & 24,850 & 24,850 & Writer & 32 & \xmark \\

    \rowcolor{row-pink}
    Social IQA~\cite{sap2019socialiqa} & 2019 & 37,588 & 37,588 & Reader & OV & \xmark \\

    \rowcolor{row-pink}
    Crowd-enVENT~\cite{troiano2023dimensional} & 2023 & 6,591 & 11,091 & Writer + Reader & 13 & \xmark\\

    \midrule
    \rowcolor{row-blue}
    \multicolumn{7}{l}{{\textit{\textbf{Cross-domain Integration (Ours)}}}} \\
    
    \rowcolor{row-blue}
    Persona-E\textsuperscript{2} & 2026 & 3,111 & 111,996 & Reader & 7 & \cmark  \\
    \bottomrule
    
  \end{tabular}}
  \caption{A unified comparison of emotion-annotated datasets across three sources: News, Social Media, and Life Experience. \textit{Note:} OV: Open vocabulary.}  
  \label{tab:DatasetCompare}
\end{table*}

\section{Related Work}
\label{sec:RelatedWork}
Extended discussions are provided in Appendix~\ref{appendix:Broad_Related_Work}.

\subsection{Event-Elicited Emotion Analysis}
Early research established the baseline for understanding emotions elicited by events. Classic works like ISEAR~\cite{scherer1994evidence}, SocialIQA~\cite{sap2019socialiqa} and others~\cite{rashkin2018event2mind, troiano2019crowdsourcing, forbes2020social} analyzed first-person narratives and social commonsense, treating events as primitive stimuli for affective responses. Subsequent studies introduced appraisal theory to interpret these cognitive layers in depth~\cite{troiano2022x, troiano2023dimensional}. Crucially, the field is shifting from writer-expressed sentiment to reader-based perception~\cite{buechel2017readers}. Benchmarks such as GoodNewsEveryone~\cite{oberlander2020goodnewseveryone}, iNews~\cite{hu2025inews} and RESEMO~\cite{hu2024resemo} focus on how audiences react to news and social media. However, most existing resources rely on aggregating annotations into a single ground truth, which obscures the inter-individual variability essential for understanding diverse emotional elicitation~\cite{plank2022problem, soni2024comparing}.

\subsection{Personality-Conditioned Affective Computing}
Research on personality–emotion interaction typically utilizes the MBTI~\cite{myers1962myers} and the BFI~\cite{john2010handbook} via three paradigms. Explicit methods link self-reported traits to text or dialogue, as seen in datasets like PANDORA~\cite{jurkovic2021pandora}, and PersonaTAB~\cite{inoue2025personatab}, though they primarily capture writer expression rather than reader elicitation. Implicit methods infer traits from behavioral data but often lack ground truth~\cite{gao2013improving, wang2024emotion, hu2024llm, shen2025emoperso}. Recently, LLM-based simulation has emerged to generate persona-specific responses~\cite{tseng2024two}, such as Big5-Chat~\cite{li2025big5}, PersonaGym~\cite{samuel2024personagym}, and PersonalityEdit~\cite{mao2024editing}. Studies show that as richer prompts with profiles are introduced, the behavioral fidelity of simulated agents improves accordingly~\cite{bai2025scaling, hu2024quantifying}. Despite their promise, recent works indicate a ``personality illusion``~\cite{han2025personality} where models mimic linguistic styles without adopting the underlying appraisal mechanisms. This highlights a critical gap: the lack of a human-grounded dataset to rigorously evaluate whether LLMs truly capture trait-driven emotional diversity.

\begin{figure*}[t]
    \centering
    \includegraphics[width=\textwidth]{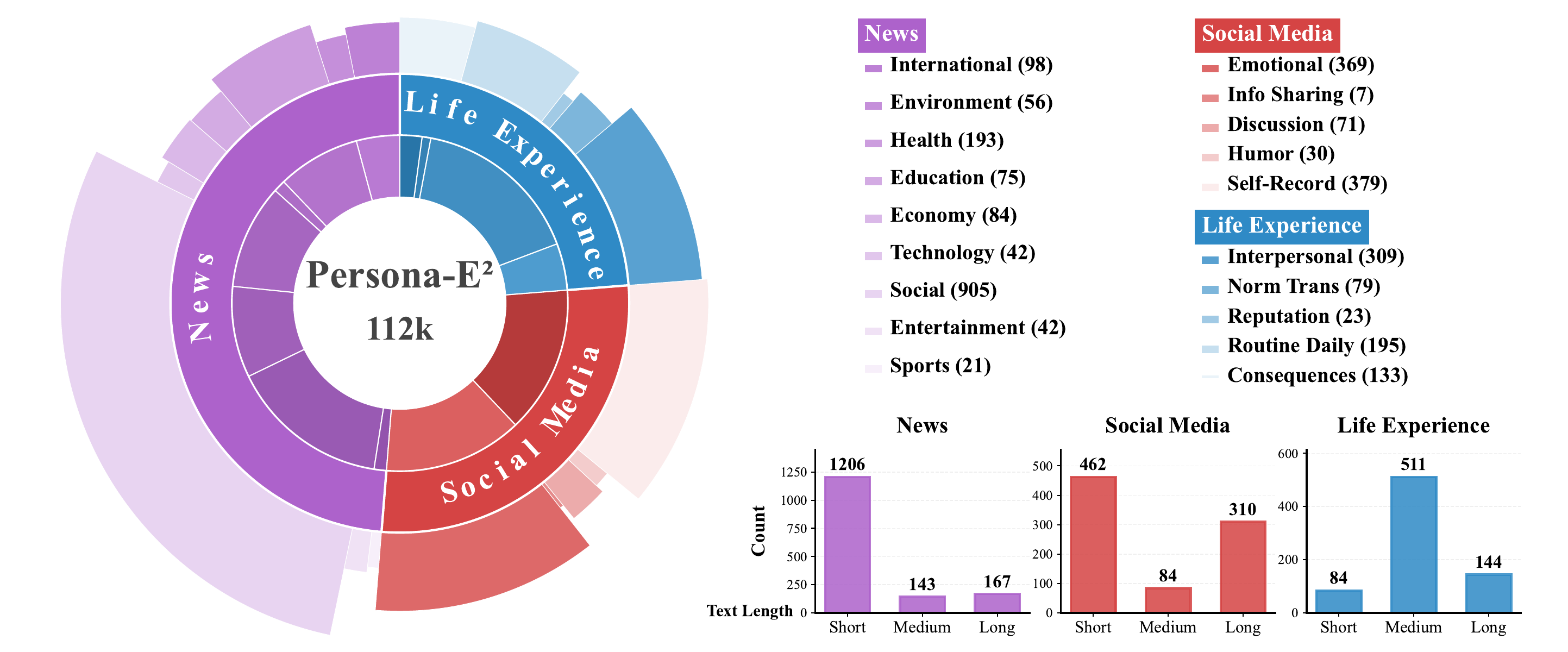} 
 \caption{Hierarchical composition of the Persona-E\textsuperscript{2} dataset. The sunburst chart illustrates three levels from inner to outer: original data sources, three primary domains, and fine-grained semantic subcategories. \textit{Note:} Text Length: Short (0--30), Medium (30--100), Long (100--300).}
    \label{fig:dataset_overall}
\end{figure*}

\section{Persona-E$^2$ Dataset Construction}
\label{sec:DatasetConstruction}

To construct a rigorously controlled dataset for reader-based emotion elicitation, we designed a pipeline integrating heterogeneous event sourcing and a multi-stage filtering process. 

\subsection{Event Sources}
To ensure affective variety and broad coverage, we gather events from three complementary domains—news, social media, and life experience narratives—covering both digital-world and real-world contexts (Appendix~\ref{appendix:CollectionSource}). These sources include two distinct elicitation modes: a) First-person projection, where personal experience drives affective memory, and b) Third-person observation, involving detached, personality-shaped appraisals. This constitutes a large-scale reader-centered emotion dataset that integrates the diversity of source domains and elicitation modes.

\paragraph{News}
We crawled factual reports from mainstream news websites, trending topics and verified institutional accounts. These well-structured texts provide socially significant events that elicit emotions from third-person perspective observations.

\paragraph{Social Media}
We collected posts from several public channels on Reddit. Social content brings greater topical breadth and more ambiguous context, which may elicit empathy or judgment from annotators. To capture more social norms and interpersonal dynamics, we also introduced a subset of events from Social Chemistry 101~\cite{forbes2020social} to ensure sufficient emotion-eliciting stimuli.

\paragraph{Life Experience}
Life experience narratives were obtained from specific channels dedicated to experience sharing. These events focus on the quotidian experiences of everyday life, ranging from minor frustrations to moments of gratitude. Such narratives are designed to invite first-person projection, serving as a counterpart to the detached perspective of news.

\subsection{Event Filtering Pipeline}

Raw collections inevitably contain noise, safety risks, and large quantities of content that lack emotional significance. To ensure high-quality stimuli, we implemented a 3-stage filtering procedure.

\paragraph{Stage 1: Content Safety Filtering.}
For life experience and social media domains, we first prune toxic or sensitive content using NSFW classifiers~\cite{albouzidi2023distilbertNSFW, tostai2023nsfwLarge}. More detailed information is illustrated in Appendix~\ref{appendix:nsfw_filter}.

\paragraph{Stage 2: Multi-Dimensional LLM Scoring.}
We utilize \llm{Qwen3-MAX}~\cite{qwen3max} to translate non-English materials into English and filter out events that lack emotional significance. The final weighted score is computed as:
\begin{equation}
\text{Score} = 0.35 V + 0.30 A + 0.20 R + 0.15 I
\end{equation}
Here, $V$, $A$, $I$, and $R$ represent personality variability, emotional arousal, emotional implicitness, and source relevance, respectively (see Appendix~\ref{appendix:Multi-Dimensional LLM Scoring}). Applying source-specific thresholds (Appendix~\ref{appendix:source_thres}), we selected 6,348 candidates from an initial pool of 76,773 events.

\paragraph{Stage 3: Expert Verification.}
Finally, a 5-member expert panel conducted a rigorous audit—removing factual errors, translation bias, and hate speech—yielding a set of 3,111 events. English examples are provided in Appendix~\ref{appendix:Dataset_Examples}.

\subsection{Annotation Protocol}
As shown in Tab.~\ref{tab:DatasetCompare}, we adopted Ekman’s six emotions (disgust, fear, anger, sadness, surprise, joy) plus neutral~\cite{ekman1992argument}. Following the ISEAR~\cite{scherer1994evidence}, we used a reader-centric question: \textit{``How would you feel when reading this event?"}, to capture elicitation rather than semantics. During this process, no role-playing was involved for human annotators so that each data point is anchored in an authentic persona.

\paragraph{Annotation Unit} We define an annotation unit as a reader-centric emotional response to a single textual event conditioned on a specific personality profile. Each unit is a tuple integrating the event context, a unique annotator ID, the annotator's measured trait scores, and the corresponding emotion label.

\paragraph{Labeling Process} 
As shown in Tab.~\ref{tab:DatasetCompare}, we adopted Ekman’s six emotions (disgust, fear, anger, sadness, surprise, joy) plus neutral~\cite{ekman1992argument}. Following the ISEAR~\cite{scherer1994evidence}, we adopted the question style: \textit{``How would you feel when reading this event?"}, to capture elicitation rather than semantics. 

\paragraph{Annotator Recruitment.}
We recruited 36 annotators, profiling their personalities via MBTI~\cite{myers1962myers} and BFI~\cite{john2010handbook} questionnaires (Appendix~\ref{sec:Annotator_Profiles}).  Crucially, the annotation process involved no role-playing. Consequently, annotators were strictly instructed to report  their genuine emotional reactions for each of the 3111 events.

\paragraph{Quality Control.}
To ensure high-fidelity responses, we implemented reader-centric training, mandatory guideline review and behavior monitoring (Appendix~\ref{sec:Quality_Control}). The annotation task was distributed over several weeks to maintain annotator attention and label quality.

\begin{figure*}[t]
    \centering
    \includegraphics[width=\textwidth]{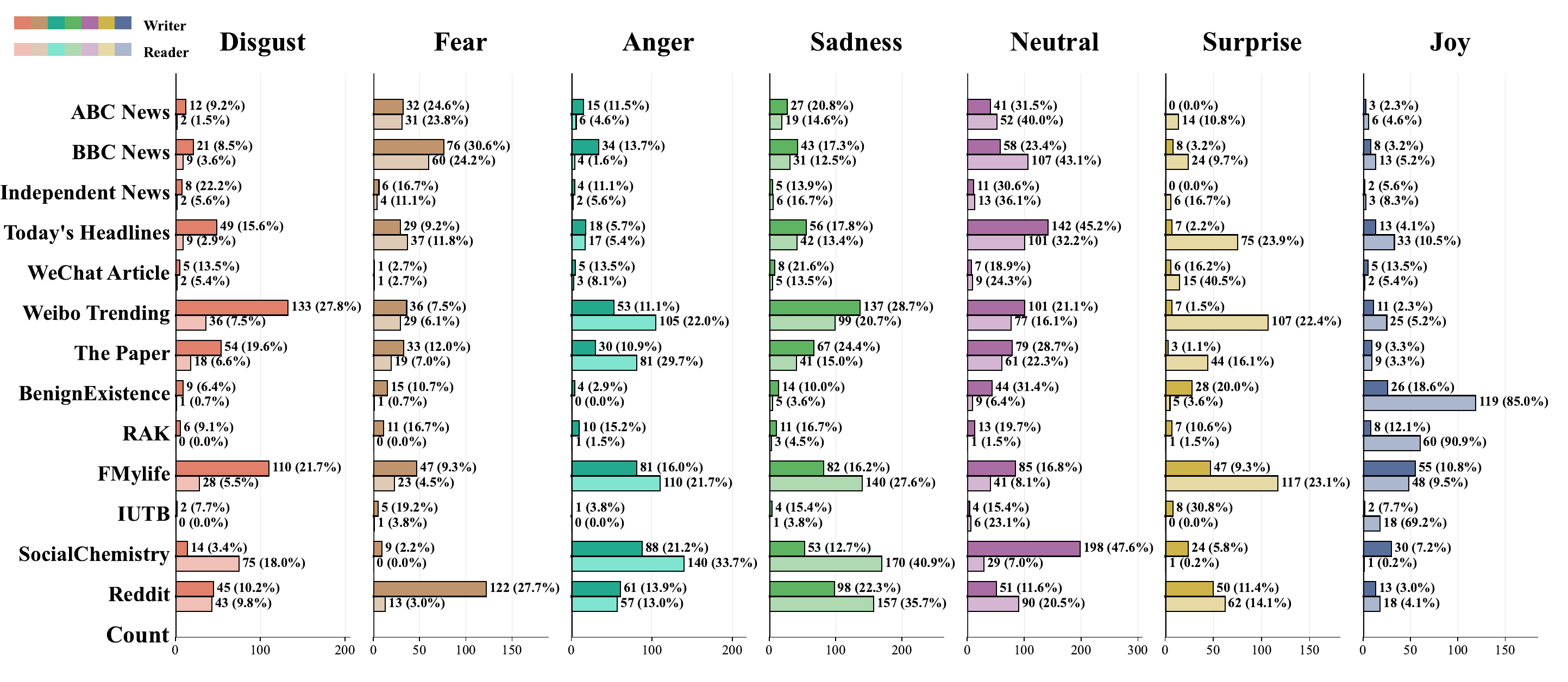} 
\caption{Comparison of emotion distributions between Writers and Readers showing the percentage distribution of seven emotions across sources . \textit{Note:} IUTB, ASK stand for r/iusedtobelieve and RandomActsofKindness.}
    \label{fig:emotion_domain}
\end{figure*}

\section{Dataset Analysis}
\label{sec:DatasetAnalysis}
\subsection{Descriptive Analysis}
Persona-E\textsuperscript{2} comprises 112k high-quality annotations across three domains: news (49\%), social media (27\%), and life experiences (24\%). As shown in Fig.~\ref{fig:emotion_domain}, a significant emotional divergence exists between the writers' sentiment~\cite{hartmann2022emotionenglish} and the readers' actual emotions. This divergence is domain-dependent: slight in factual news but pronounced in life experiences, where interpersonal narratives trigger first-person projection and emotional transmission. Moreover, the observed shift from writer to reader emotion leads us to a deeper discussion of \textbf{RQ1} (Sec.~\ref{sec:Dataset_Affective_Divergence}).

\subsection{Annotation Reliability}
In affective computing, inter-annotator disagreement is often dismissed as label noise~\cite{plank2022problem}. Moving beyond this, we report the annotation reliability by testing the personality-aware agreement gap. This shifts the focus from universal consensus to trait-conditioned alignment, grounded in the premise that subjective disagreement is structured by latent personality profiles. Thus, annotators with similar personalities should exhibit higher consensus than random groupings. The validation of this hypothesis via BFI and MBTI clustering ensures the dataset's reliability.

\paragraph{In-Group Agreement}
To validate the hypothesis, we applied K-means clustering on BFI vectors. Unlike MBTI's discrete categories, this method adapts to continuous traits and ensures balanced clusters, addressing the statistical instability that arises from MBTI's sparse subgroups.  The choice of $k=6$ was empirical, aimed at balancing the number of annotators per cluster with the captured personality diversity. To ensure the robustness of our groupings, we conducted a sensitivity analysis across various algorithms (K-means, GMM, Hierarchical) and cluster counts ($k \in [3, 9]$). As detailed in Appendix \ref{appendix:Clustering_Stability}, the Personality Agreement Gap (PAG) consistently remains positive across all settings, confirming that trait-aligned grouping captures shared interpretative logic rather than clustering artifacts.

\begin{figure}[h]
    \centering
    \includegraphics[width=\linewidth]{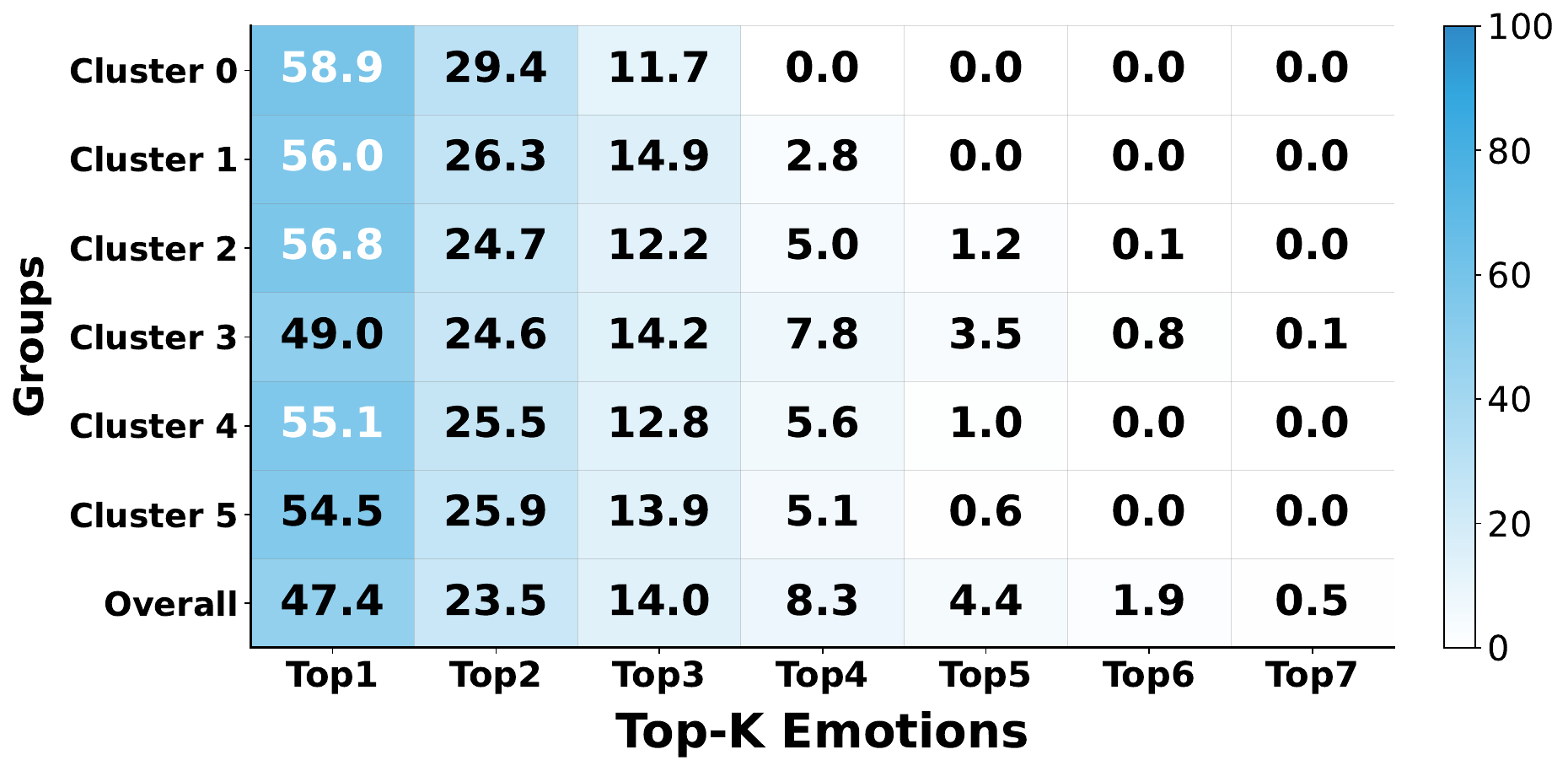}
    
    \caption{Top-$K$ emotion distribution across BFI clusters. The heatmap shows the average vote share (\%) of the $k$-th most frequent emotion per event.}
    
    \label{fig:topk_heatmap}
\end{figure}

As shown in Fig.~\ref{fig:topk_heatmap}, in-group Top-1 agreement consistently outperforms the global average. More specifically, Cluster 0 achieves a +11.5\% gain on Top-1 agreement over the baseline. This confirms that grouping by traits uncovers shared interpretative logic.

\begin{table*}[t!]
    \centering
    \small
    % --- 第一行：一致性对比 (保持不变) ---
    \begin{subtable}[h!]{0.32\linewidth}
        \centering
        \resizebox{1\linewidth}{!}{
        \begin{tabular}{lccc}
            \textbf{Cluster} & \textbf{$Agr_{in}$} & \textbf{$Agr_{out}$} & \textbf{PAG} \\
            \midrule
            \textbf{Cluster 0} & $61.07$ & $35.11$ & \underline{$+25.96$} \\
            \textbf{Cluster 1} & $57.63$ & $37.59$ & $+20.03$ \\
            \textbf{Cluster 2} & $57.26$ & $43.39$ & $+13.87$ \\
            \textbf{Cluster 3} & $49.39$ & $41.12$ & $+8.27$ \\
            \textbf{Cluster 4} & $54.73$ & $40.38$ & $+14.36$ \\
            \textbf{Cluster 5} & $54.76$ & $37.11$ & $+17.65$ \\
        \end{tabular}}
    \caption{\textbf{BFI K-means ($K=6$)}}
    \label{tab:bfi_6_consist}
    \end{subtable}
    ~
    \begin{subtable}[h!]{0.32\linewidth}
        \centering
        \resizebox{1\linewidth}{!}{
        \begin{tabular}{lccc}
            \textbf{Cluster} & \textbf{$Agr_{in}$} & \textbf{$Agr_{out}$} & \textbf{PAG} \\
            \midrule
            \textbf{Cluster 0} & $58.05$ & $37.63$ & \underline{$+20.42$} \\
            \textbf{Cluster 1} & $56.32$ & $38.13$ & $+18.19$ \\ 
            \textbf{Cluster 2} & $51.05$ & $45.15$ & $+5.9$ \\ 
            \textbf{Cluster 3} & $49.39$ & $41.37$ & $+8.02$ \\ 
            ~ & ~ & ~ & ~ \\
            ~ & ~ & ~ & ~ \\
        \end{tabular}}
    \caption{\textbf{BFI K-means ($K=4$)}}
    \label{tab:bfi_4_consist}
    \end{subtable}
    ~
    \begin{subtable}[h!]{0.30\linewidth}
        \centering
        \resizebox{1\linewidth}{!}{
        \begin{tabular}{lccc}
            \textbf{Type} & \textbf{$Agr_{in}$} & \textbf{$Agr_{out}$} & \textbf{PAG} \\
            \midrule
            \textbf{ESTP} & $64.33$ & $37.35$ & \underline{$+26.98$} \\
            \textbf{INTP} & $60.03$ & $36.83$ & $+23.20$ \\
            \textbf{INTJ} & $61.53$ & $37.58$ & $+23.95$ \\
            \textbf{ESTJ} & $56.02$ & $38.39$ & $+17.63$ \\
            \textbf{ENTJ} & $55.30$ & $35.84$ & $+19.46$ \\
            \textbf{ISTJ} & $50.83$ & $41.16$ & $+9.68$ \\
        \end{tabular}}
    \caption{\textbf{MBTI Grouping ($N \ge 3$)}}
    \label{tab:mbti_consist}
    \end{subtable}

    % --- 第二行：BFI均值特征 (添加下划线与缩写解释) ---
    \begin{subtable}[h!]{0.48\linewidth}
        \centering
        \resizebox{1\linewidth}{!}{
        \begin{tabular}{lcccccc}
            \textbf{Group} & \textbf{$N$} & \textbf{Open.} & \textbf{Cons.} & \textbf{Extra.} & \textbf{Agree.} & \textbf{Neuro.} \\
            \midrule
            \textbf{Cluster 0} & 3 & \underline{$0.823$} & \underline{$0.868$} & \underline{$0.691$} & $0.622$ & $0.358$ \\
            \textbf{Cluster 1} & 4 & $0.718$ & $0.536$ & $0.482$ & $0.797$ & $0.638$ \\
            \textbf{Cluster 2} & 7 & $0.553$ & $0.647$ & $0.629$ & $0.673$ & $0.435$ \\
            \textbf{Cluster 3} & 11 & $0.695$ & $0.828$ & $0.604$ & \underline{$0.801$} & $0.206$ \\
            \textbf{Cluster 4} & 6 & $0.556$ & $0.569$ & $0.455$ & $0.580$ & \underline{$0.592$} \\
            \textbf{Cluster 5} & 5 & $0.612$ & $0.763$ & $0.494$ & $0.673$ & $0.335$ \\
        \end{tabular}}
    \caption{\textbf{BFI Mean Features ($K=6$)}}
    \label{tab:bfi_6_traits}
    \end{subtable}
    ~
    \begin{subtable}[h!]{0.48\linewidth}
        \centering
        \resizebox{1\linewidth}{!}{
        \begin{tabular}{lcccccc}
            \textbf{Group} & \textbf{$N$} & \textbf{Open.} & \textbf{Cons.} & \textbf{Extra.} & \textbf{Agree.} & \textbf{Neuro.} \\
            \midrule
            \textbf{Cluster 0} & 4 & \underline{$0.815$} & \underline{$0.836$} & \underline{$0.628$} & $0.636$ & $0.336$ \\
            \textbf{Cluster 1} & 5 & $0.679$ & $0.517$ & $0.429$ & $0.629$ & \underline{$0.729$} \\
            \textbf{Cluster 2} & 16 & $0.560$ & $0.661$ & $0.559$ & $0.682$ & $0.432$ \\
            \textbf{Cluster 3} & 11 & $0.695$ & $0.828$ & $0.604$ & \underline{$0.801$} & $0.206$ \\
            ~ & ~ & ~ & ~ & ~ & ~ & ~ \\
            ~ & ~ & ~ & ~ & ~ & ~ & ~ \\
        \end{tabular}}
    \caption{\textbf{BFI Mean Features ($K=4$)}}
    \label{tab:bfi_4_traits}
    \end{subtable}

    \caption{Personality-based Top-1 agreement analysis. \textbf{Top:} Comparison of In-Group ($Agr_{in}$) and Out-Group ($Agr_{out}$) agreement levels ($N$ controlled). \textbf{Bottom:} Mean BFI traits for clusters. \textit{Note:} Open., Cons., Extra., Agree., and Neuro. represent the Big Five traits; $N$ denotes the number of annotators.}
    \label{tab:personality_full_analysis}
\end{table*}

\paragraph{Personality Grouping Analysis}

As shown in Tab.~\ref{tab:personality_full_analysis}, we quantify the personality effect by calculating the \textbf{Personality Agreement Gap} ($PAG = Agr_{in} - Agr_{out}$), representing the Top-1 agreement delta between In-Group and Out-Group pairs with equalized sample sizes. 
\begin{itemize}
\item \textbf{BFI Grouping:} Cluster 0 (High Conscientiousness/Openness) shows a massive PAG of +25.96\% compared to the out-group (Tab.~\ref{tab:bfi_6_consist}), indicating a convergent appraisal pattern. Conversely, Cluster 3 (Low Neuroticism) exhibits the smallest PAG (+8.3\%). This may suggest that traits like high Neuroticism act as strict ``perceptual filters'' that funnel reactions into specific categories. Consistent patterns are also revealed across different $K$ settings (Tab.~\ref{tab:bfi_4_consist}). 
\item \textbf{MBTI Grouping:} Similar patterns appear in MBTI types (Tab.~\ref{tab:mbti_consist}). ESTP types achieve peak PAG (+26.98\%) by prioritizing social cues~\cite{pickett2004getting}, while ISTJ yields lower consensus.
\end{itemize}
Positive PAG values reveal that in-group agreement consistently exceeds out-group agreement, providing empirical evidence that affective disagreement is structured by trait-aligned patterns.

\section{Experiment}
\label{sec:Experiment}
\subsection{Experimental Setup}
We designed experiments to assess the necessity of personality modeling in emotion analysis and the capability of LLMs to simulate these affective shifts. To evaluate performance for \textbf{RQ2} and \textbf{RQ3}, we employed leading open and closed-source models.

\paragraph{Model List:} \llm{GPT-5.1}~\cite{openai2025gpt51}, \llm{Llama-3-8B}~\cite{dubey2024llama}, \llm{Qwen3-8B}~\cite{yang2025qwen3}, \llm{Gemma-3-12B}~\cite{team2025gemma}, and \llm{Ministral-3-8B}~\cite{mistral2025ministral3hf}.
\paragraph{Computing Environment:} Open-source models are evaluated on a cloud computing platform using NVIDIA A100 GPUs. Closed-source models are accessed through their APIs. Appendix~\ref{appendix:Implementation_Details} presents details on model version and hyper-parameters.

\subsection{RQ1. Dataset Affective Divergence}
\label{sec:Dataset_Affective_Divergence}

\begin{figure*}[t!] % 跨栏置顶
    \centering
    
    % --- 第一张图 (左) ---
    % 将宽度设为 0.32\textwidth (约1/3)，预留一点间隙
    \begin{subfigure}{0.32\textwidth} % 0.36
        \centering
        % 建议加上 keepaspectratio，虽然通常默认就是如此
        \includegraphics[width=\linewidth, height=3.0cm]{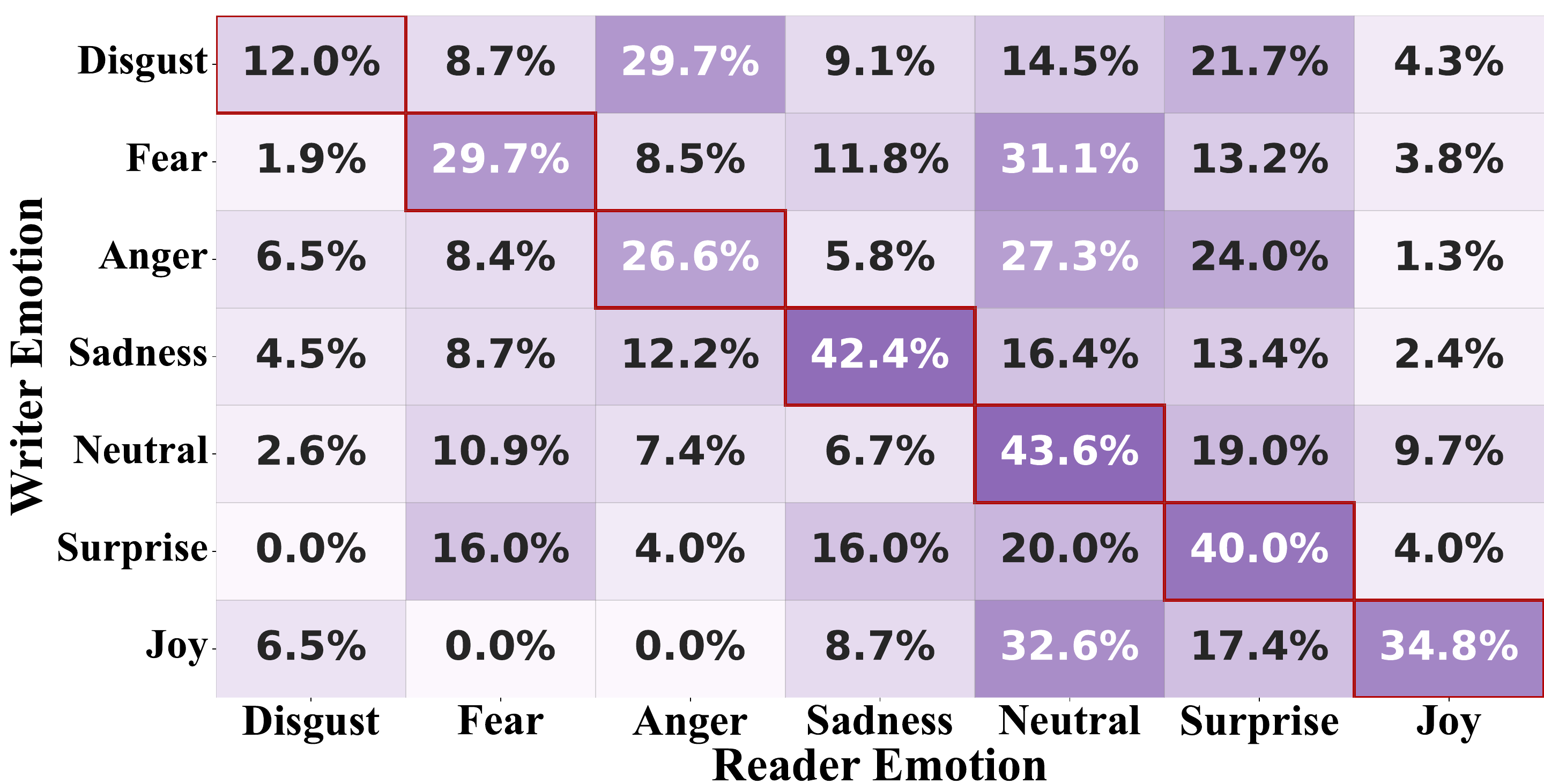}
        \caption{News}
        \label{fig:trans_w2r_news}
    \end{subfigure}
    \hfill % \hfill 用于在图之间自动填充空白，使分布均匀
    % --- 第二张图 (中) ---
    \begin{subfigure}{0.32\textwidth} % 0.31
        \centering
        \includegraphics[width=\linewidth, height=3.0cm]{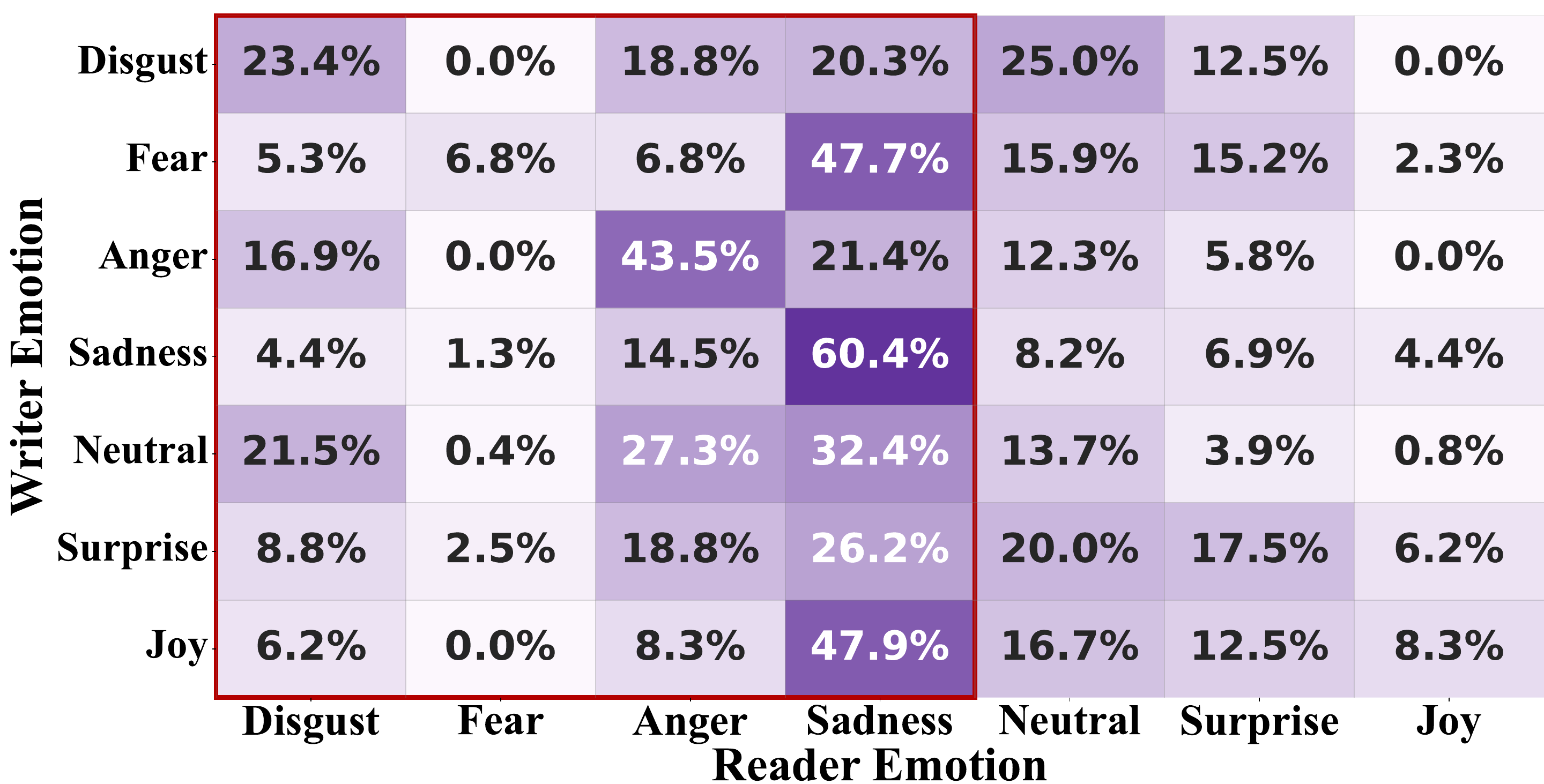}
        \caption{Social Media}
        \label{fig:trans_w2r_social}
    \end{subfigure}
    \hfill % \hfill
    % --- 第三张图 (右) ---
    \begin{subfigure}{0.34\textwidth} % 0.31
        \centering
        \includegraphics[width=\linewidth, height=3.0cm]{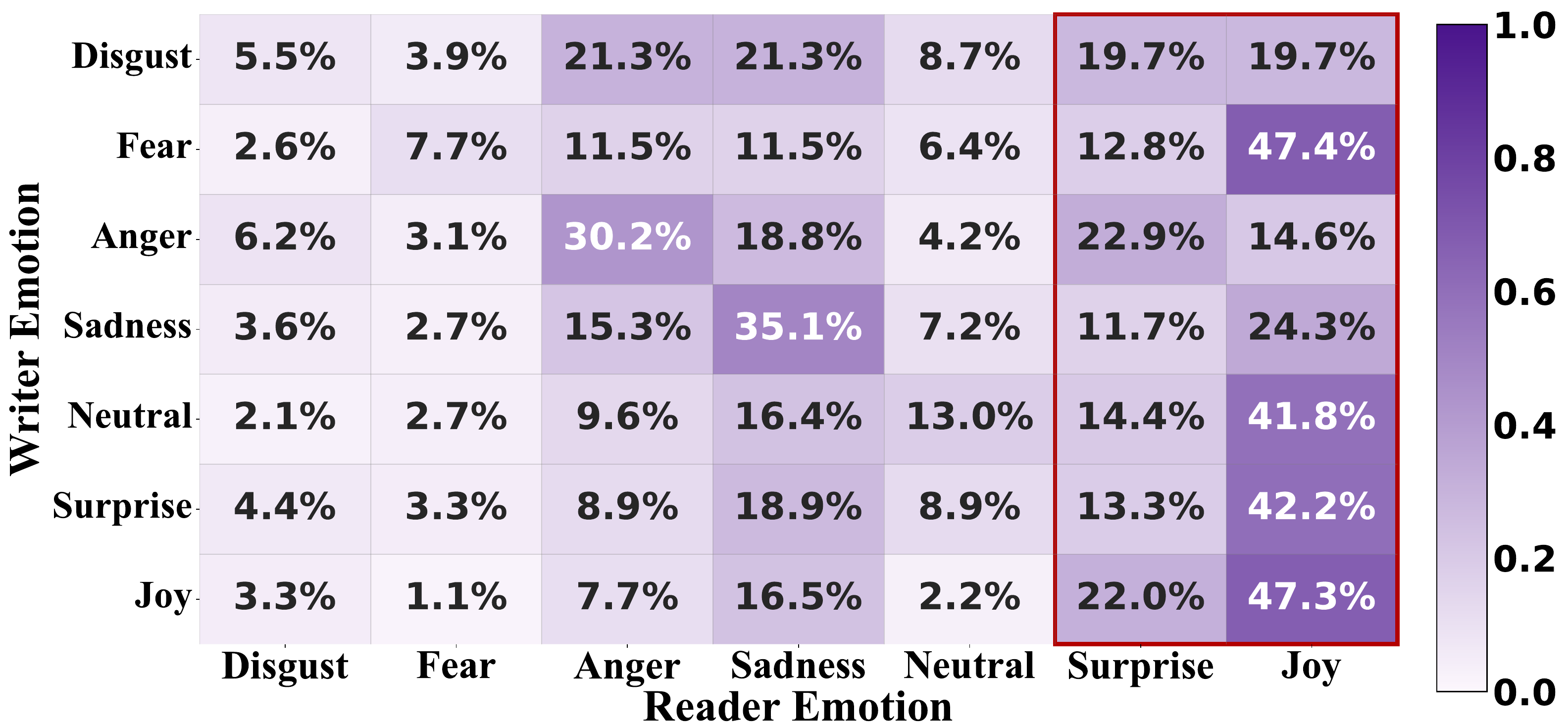}
        \caption{Life Experience}
        \label{fig:trans_w2r_life}
    \end{subfigure}
    % --- 主标题 ---
    \caption{Affective transition matrices between General Writer and Reader. Red boxes highlight: (a) high resonance in News, (b) negative bias in Social Media, and (c) positive shift in Life Experience. \textit{Note:} Negative: disgust, fear, anger, sadness; Positive: surprise, joy.}
    \label{fig:three_heatmaps_row}
\end{figure*}

\textbf{How do emotional responses diverge across the General Writer, General Reader, and Persona Reader?} Fig.~\ref{fig:emotion_domain} illustrates a gap between writer and reader-based emotions, which we hypothesize is modulated by domain and personality-driven appraisal patterns~\cite{buechel2017readers}. To validate the assumption, we differentiated three affective layers: (1) General Writer (GW): semantic sentiment predicted by a pre-trained emotion classifier~\cite{hartmann2022emotionenglish} that provides an identical label space to our human annotations; (2) General Reader (GR): majority-vote elicitation; and (3) Persona Reader (PR): trait-conditioned elicitation. To quantify this divergence, we categorized seven emotions into positive (surprise, joy) and negative (disgust, fear, anger, sadness) polarities. We then computed affective transition matrices, where element $(i, j)$ represents the probability of emotion $i$ shifting to emotion $j$ across the three perspectives.

\paragraph{Domain-Driven Divergence.} 
Comparing semantic sentiment (GW) with majority-vote elicitation (GR) reveals that emotional appraisal is a domain-sensitive reconstruction rather than a direct transfer (Fig.~\ref{fig:three_heatmaps_row}). Appendix~\ref{appendix:rq1.1} presents the detailed data analysis related to this finding.
First, news acts as a rational buffer: it maintains a high neutrality transfer and emotional resonance rate (Tab.~\ref{tab:rq1.1_polar1}). This aligns with professional journalism's role in minimizing cognitive appraisal variance through factual grounding~\cite{1991Emotion, alm2008affect}. Second, social media functions as an ``Emotional Black Hole''~\cite{baumeister2001bad}. We observed a severe negativity bias, where significant portions of neutral and positive GW sentiments shift to negative GR reactions (81.6\% of neutral and 59.35\% of positive sentiments transfer into negative in Tab.~\ref{tab:rq1.1_polar2}). This reflects a ``Forced Siding'' mechanism, where ambiguity is treated as a vacuum filled by reader-side hostility~\cite{tajfel2010social, baumeister2001bad}. Conversely, life narratives trigger a ``Psychological Immune System''~\cite{gilbert1998immune}. Readers exhibit an optimism bias, filtering distress to prioritize empathetic resonance (43.3\% of negative and 56.2\% of neutral sentiments transfer to positive in Tab.~\ref{tab:rq1.1_polar3}), a cognitive nuance often overlooked by traditional sentiment analysis~\cite{matlin2016pollyanna}.

\begin{figure}[h]
    \centering
    % 第一张子图
    \begin{subfigure}{0.48\columnwidth}
        \centering
        \includegraphics[width=\linewidth]{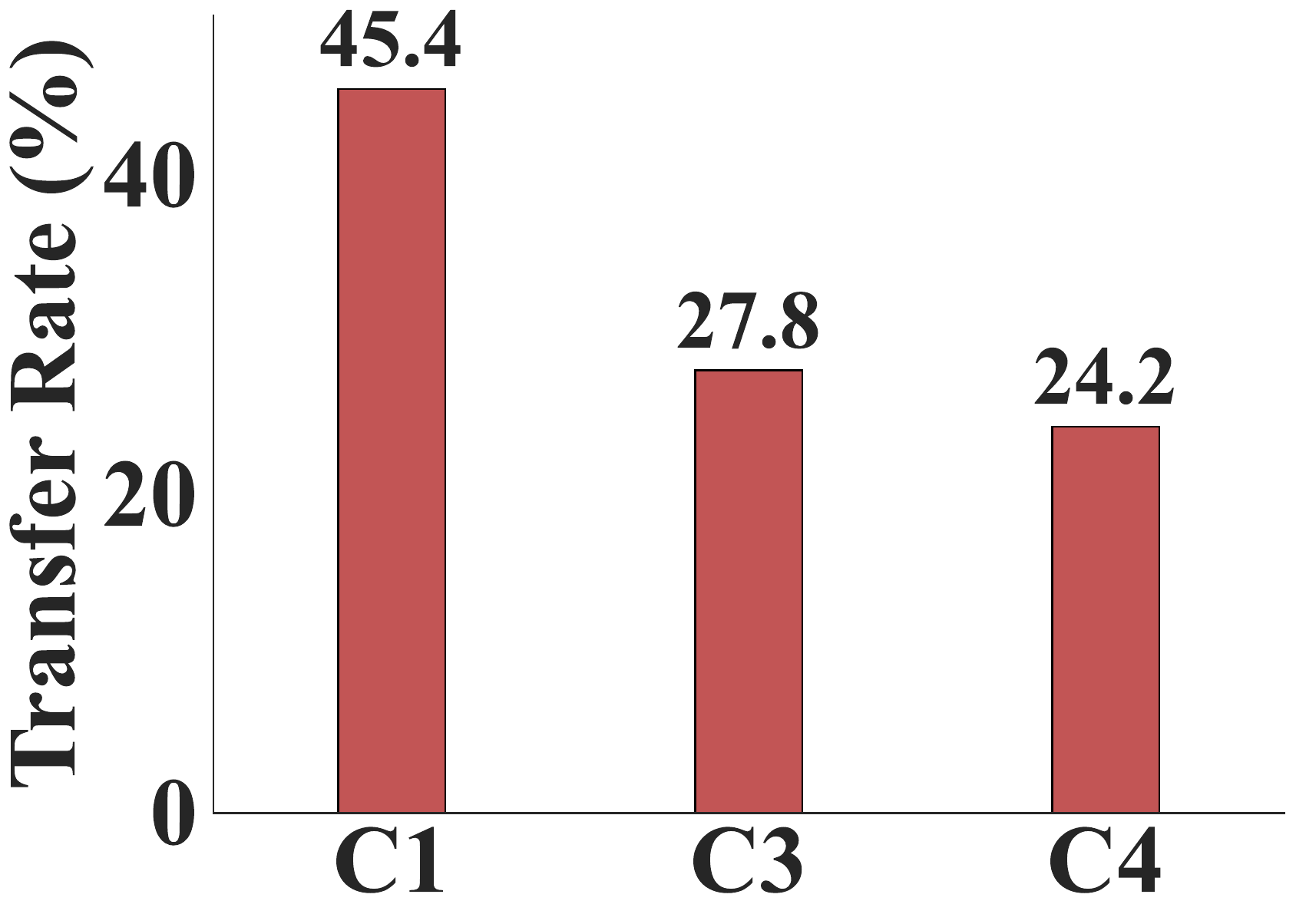} 
        \caption{Anxious Empathy}
        \label{fig:rq1.2.sub1}
    \end{subfigure}
    \hfill 
    % 第二张子图
    \begin{subfigure}{0.48\columnwidth}
        \centering
        \includegraphics[width=\linewidth]{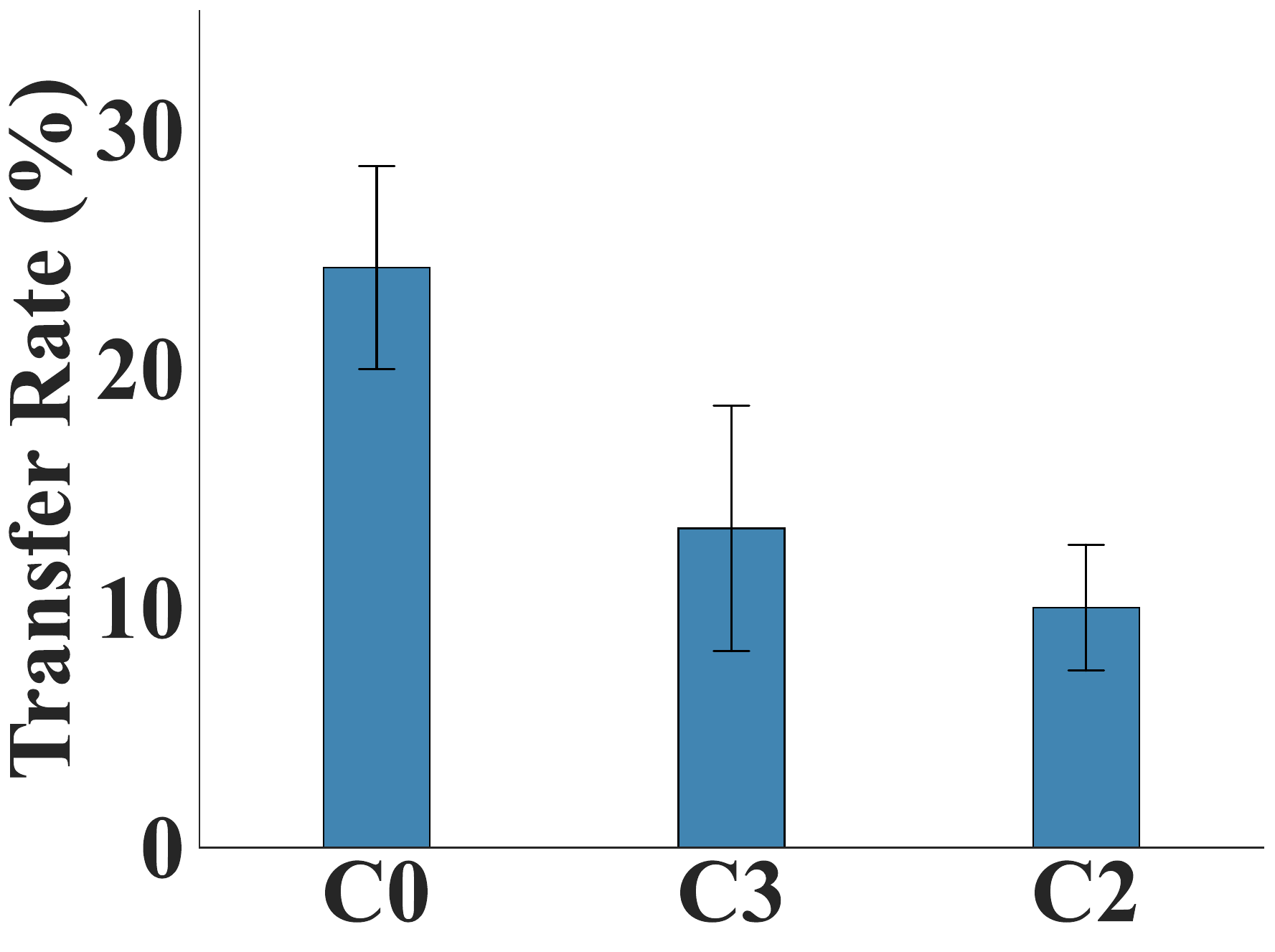}
        \caption{Negative Passivation}
        \label{fig:rq1.2.sub2}
    \end{subfigure}
    
    \caption{(a) Transfer rate from non-negativity to negativity in the news domain, and (b) Transfer rate from negativity to non-negativity, error bar means deviation among 3 domains.}
\end{figure}

\paragraph{Personality-Driven Modulation.} 
We employed BFI-based clustering for analysis rather than MBTI categories, as the latter led to sparse populations in specific groups, which cannot provide reliable statistics. We found that personality significantly acts as a filter for emotional transfer (further data analysis is available in Appendix~\ref{sec:rq1_r2r}). First, Anxious Empathy: Individuals with high Agreeableness (A) and Neuroticism (N), such as C1, show the highest neutral-to-negative transfer rates. Here, high A's social sensitivity is likely hijacked by the interpretation bias of high N~\cite{mccrae1997conceptions}, as evidenced in Fig.~\ref{fig:rq1.2.sub1}. Secondly, Negativity Passivation: High Conscientiousness (C) and Extraversion (E) exhibit ``Negative Passivation.'' C0 shows the lowest negative resonance in Fig~\ref{fig:rq1.2.sub2}. This suggests these individuals effectively regulate distress~\cite{gross1998emerging}. In contrast, low C/E (e.g., C2 and C3) led to ``Negative Locking'' due to a lack of perceived control. Finally, Neutralization: Illustrated in Tab.~\ref{tab:rq1_r2r_insight_c}, we found a statistical correlation between Openness (O) and the Neutralization Rate ($r=+0.86$, $p=0.027$). To align with cognitive theory, high-O individuals use cognitive complexity to moderate emotional activation, showing a low need for cognitive closure~\cite{webster1994individual}.
In summary, emotional elicitation is reshaped by domain effects and personality traits, highlighting the move from general sentiment analysis to persona-aware modeling.

% RQ2 结果
% 颜色定义
\definecolor{color1}{HTML}{EBF5EE} % 对应 Task Group 1
\definecolor{color2}{HTML}{E8F0FE} % 对应 Task Group 2
\definecolor{color3}{HTML}{F3E5F5} % 对应 Task Group 3
\definecolor{color4}{HTML}{EFEBE9} % 对应 Task Group 4 (原MDS颜色)

% 自定义列类型：固定宽度并自动填充背景色
% 为了能在有限的页面宽度内放下13列，将宽度设为约 1.1cm
\newcolumntype{A}{>{\columncolor{color1}\centering\arraybackslash}p{1.1cm}}
\newcolumntype{B}{>{\columncolor{color2}\centering\arraybackslash}p{1.1cm}}
\newcolumntype{C}{>{\columncolor{color3}\centering\arraybackslash}p{1.1cm}}
\newcolumntype{D}{>{\columncolor{color4}\centering\arraybackslash}p{1.1cm}}

\begin{table*}[t!]
  \centering
  \footnotesize
  
  % --- 压缩行高与消除缝隙 ---
  % \setlength{\aboverulesep}{0pt}
  % \setlength{\belowrulesep}{0pt}
  % \renewcommand{\arraystretch}{0.95}
  \setlength\tabcolsep{1.5pt} % 稍微减小间距以容纳新增的列
  \resizebox{\textwidth}{!}{%
      \begin{tabular}{ll AAA BBB CCC DDD}
        \toprule
        % 第一行表头，注意 multicolumn 现在从第3列开始跨度
        \multirow{2}{*}{\rotatebox{90}{\textbf{Metric}}}  & \textbf{Source} & \multicolumn{3}{c}{\textbf{News}} & \multicolumn{3}{c}{\textbf{Social Media}} & \multicolumn{3}{c}{\textbf{Life Experience}} & \multicolumn{3}{c}{\textbf{Overall}} \\ 
        \cmidrule(lr){3-5} \cmidrule(lr){6-8} \cmidrule(lr){9-11} \cmidrule(lr){12-14}
        
        % 第二行表头
        & \textbf{Prompt} & 
        \multicolumn{1}{c}{\textbf{General}} & \multicolumn{1}{c}{\textbf{Persona}} & \multicolumn{1}{c}{\textbf{CoT}} & 
        \multicolumn{1}{c}{\textbf{General}} & \multicolumn{1}{c}{\textbf{Persona}} & \multicolumn{1}{c}{\textbf{CoT}} & 
        \multicolumn{1}{c}{\textbf{General}} & \multicolumn{1}{c}{\textbf{Persona}} & \multicolumn{1}{c}{\textbf{CoT}} & 
        \multicolumn{1}{c}{\textbf{General}} & \multicolumn{1}{c}{\textbf{Persona}} & \multicolumn{1}{c}{\textbf{CoT}} \\
        \midrule
        
        % --- Top-1 Accuracy 部分 ---
        \multirow{5}{*}{\rotatebox{90}{\textbf{Top-1 Acc.}}} 
        & \llm{GPT5.1}          & \textbf{31.8} & 25.0 & 29.5 & 18.2 & 18.2 & \textbf{27.3} & 32.4 & 35.3 & 35.3 & 29.0 & 27.0 & \textbf{31.0} \\
        & \llm{Llama3-8B}       & 29.5 & 29.3 & 20.5 & 9.1 & 4.8 & 13.6 & 32.4 & 32.4 & 32.4 & 26.0 & 25.0 & 23.0 \\
        & \llm{Qwen3-8B}        & 18.2 & 26.9 & 25.4 & 23.1 & 21.4 & 22.7 & 20.3 & 33.7 & 26.0 & 20.0 & 28.0 & 25.0 \\
        & \llm{Gemma3-12B}      & 18.2 & 25.0 & 27.3 & 22.7 & \textbf{27.3} & 18.2 & 35.3 & 32.4 & 29.4 & 25.0 & 28.0 & 26.0 \\
        & \llm{Ministral3-8B}   & 18.2 & 13.6 & 20.5 & 5.0 & 9.1 & 4.5 & \textbf{36.4} & 35.3 & 35.3 & 22.0 & 20.0 & 22.0 \\
        
        \midrule % 中间分割线
        
        % --- Top-2 Accuracy 部分 ---
        \multirow{5}{*}{\rotatebox{90}{\textbf{Top-2 Acc.}}} 
        & \llm{GPT5.1}          & 54.5 & 59.1 & \textbf{59.1} & 40.9 & \textbf{50.0} & 45.5 & 47.1 & 55.9 & \textbf{55.9} & 49.0 & \textbf{56.0} & 55.0 \\
        & \llm{Llama3-8B}       & 47.7 & 43.9 & 45.5 & 18.2 & 19.0 & 31.8 & 50.0 & 47.1 & 44.1 & 42.0 & 39.6 & 42.0 \\
        & \llm{Qwen3-8B}        & 31.3 & 33.5 & 38.6 & 31.4 & 36.4 & 39.4 & 39.2 & 39.0 & 45.1 & 34.0 & 36.0 & 41.0 \\
        & \llm{Gemma3-12B}      & 36.4 & 47.7 & 56.8 & 27.3 & 40.9 & 45.5 & 50.0 & 52.9 & 41.2 & 39.0 & 48.0 & 49.0 \\
        & \llm{Ministral3-8B}   & 40.9 & 29.5 & 38.6 & 35.0 & 36.4 & 22.7 & 51.5 & 50.0 & 50.0 & 43.3 & 38.0 & 39.0 \\  
        \bottomrule
      \end{tabular}%
      }
      \caption{Comparison of different models performance in subset. \textit{Note:}  General Prompt: Prompt without personality traits, Persona: Prompt with BFI traits, CoT: CoT-Style Prompt with BFI traits.}
      \label{tab:hard_set_results}
\end{table*}

\subsection{RQ2. LLM Emotion Simulation}
\label{sec:LLM_Emotion_Prediction}
\textbf{Can LLMs effectively mimic personality-shaped emotional responses?} We investigated whether models can predict emotional shifts with personality profiles. LLMs are conditioned on three strategies: General Prompt, Persona Prompt (BFI personality vectors~\cite{JOHNSON201478}), and Persona-CoT (Chain of Thought)~\cite{wei2022chain}.

While general performance on 100 randomly sampled events is reported in Tab.~\ref{tab:rq2_general_results}, we specifically focus on trait-driven shifts by constructing the Subjective Divergence Subset (SDS). SDS targets scenarios where emotional responses are clear but heavily conditioned on the annotator's personality. Label validity is ensured via the \textit{Group Consensus Score} ($S_{consensus}$). For a probability distribution $P_G$ of group $G$, consensus is defined as:

\begin{equation}
    S_{consensus}(G) = 1 - \frac{-\sum p_i \log p_i}{\log K}
\end{equation}
Thus, the SDS events satisfy $S_{consensus}(G)>\alpha$ for all personality groups~$G$, totaling 413 events (Life: 87, News: 257, Social: 69) for $\alpha=0.3$. Details of SDS are reported in Appendix~\ref{appendix:RQ2}.

\paragraph{Latent Emotion Understanding.}
As detailed in Tab.~\ref{tab:hard_set_results}, experiments on the subset reveal a significant gap between Top-1 ($\sim$25.0\%) and robust Top-2 performance ($\sim$45.0\%). This suggests that LLMs successfully map emotional contexts to relevant semantic neighborhoods~\cite{NEURIPS2020_1457c0d6} but fail to pinpoint the precise label. While models capture the general affective sphere, they cannot distinguish subtle sentiment shifts, indicating more human feedback is likely required in affective training mechanisms~\cite{NEURIPS2022_b1efde53}.

\paragraph{Prompt Strategies.}
We observed that persona prompts and CoT prompts are not universally beneficial; their efficacy is modulated by model capacity. Complex prompting improved the reasoning process for larger models, but degraded it in smaller architectures. For instance, \llm{GPT-5.1} increased from 29.0\% to 31.0\% but \llm{LLAMA3-8B} declined. We attribute this to attention dilution~\cite{kaplan2020scaling}, where the personality profiles may overwhelm the limited context of smaller models.

\paragraph{Domain Discrepancy.}
Models consistently struggled in the social media domain compared to others (\llm{GPT-5.1}: 18.2-27.3\% in social for Top-1). This performance gap suggests that current LLM training corpora are biased towards structured materials, failing to process the informal dynamics of virtual interactions. To bridge this gap in cyber-social emotional analysis, future studies must focus on the unstructured online materials. See Appendix~\ref{appendix:rq2_data_analysis} for a full data breakdown.

\subsection{RQ3. Cognitive Soundness}
\textbf{Do LLMs generate human-like rationales for the emotional activation?}
Beyond prediction, we evaluated the cognitive plausibility of the psychological reasoning process~\cite{zhang2023exploring, li2024evaluating} on all 413 SDS events. Five trained reviewers performed a best-of-three forced-choice evaluation~\cite{kiritchenko2016capturing} based on: a) Persona Consistency, b) Reasoning Plausibility, and c) Emotion Specificity (Definitions in Appendix~\ref{appendix:rq3}). In Tab.~\ref{tab:winrate}, selections were aggregated to compute win rates for each model.

% 首先确保在导言区定义了颜色（如果已有请忽略）
\definecolor{color7}{RGB}{235, 245, 250} % 示例浅蓝
\definecolor{color8}{RGB}{250, 240, 230} % 示例浅红
\definecolor{color9}{RGB}{240, 255, 240} % 示例浅绿

% 自定义列类型：修改为 c 类型以自适应单栏宽度
\newcolumntype{E}{>{\columncolor{color7}}c}
\newcolumntype{F}{>{\columncolor{color8}}c}
\newcolumntype{G}{>{\columncolor{color9}}c}

\begin{table}[t!] % 改为单栏 table 环境
    \centering
    % \footnotesize
    \setlength\tabcolsep{2.5pt} % 调整列间距
    % 使用 resizebox{\columnwidth} 确保表格刚好占满单栏宽度
    \resizebox{\columnwidth}{!}{%
        \begin{tabular}{l EEE FFF GGG} 
            \toprule
            % 表头第一行：Q1, Q2, Q3
            \multirow{2}{*}{\textbf{Model}} & \multicolumn{3}{c}{\textbf{Consistency}} & \multicolumn{3}{c}{\textbf{Plausibility}} & \multicolumn{3}{c}{\textbf{Specificity}} \\ 
            \cmidrule(lr){2-4} \cmidrule(lr){5-7} \cmidrule(lr){8-10}

            % 第二行表头
            & 
            \multicolumn{1}{c}{\textbf{BL}} & \multicolumn{1}{c}{\textbf{MBTI}} & \multicolumn{1}{c}{\textbf{BFI}} & 
            \multicolumn{1}{c}{\textbf{BL}} & \multicolumn{1}{c}{\textbf{MBTI}} & \multicolumn{1}{c}{\textbf{BFI}} & 
            \multicolumn{1}{c}{\textbf{BL}} & \multicolumn{1}{c}{\textbf{MBTI}} & \multicolumn{1}{c}{\textbf{BFI}} \\
            \midrule
            % 表头第二行：Gen/Per/CoT
            % & \textbf{Gen.} & \textbf{Per.} & \textbf{CoT} & 
            %   \textbf{Gen.} & \textbf{Per.} & \textbf{CoT} & 
            %   \textbf{Gen.} & \textbf{Per.} & \textbf{CoT} \\
            % \midrule
            
            % 数据行：仅保留 Top-1 Accuracy 数据
            \llm{GPT5.1}        & 4.7 & 16.5 & \textbf{78.8} & 13.8 & 15.8 & \textbf{70.4} & 17.6 & 13.5 & \textbf{68.9} \\
            \llm{Llama3}        & 9.4 & 19.8 & \textbf{70.8} & 21.5 & 23.1 & \textbf{55.4} & 20.5 & 18.2 & \textbf{61.3} \\
            \llm{Qwen3}         & 10.4 & 19.4 & \textbf{70.2} & 19.7 & 20.4 & \textbf{59.9} & 23.7 & 19.4 & \textbf{56.9} \\
            \llm{Gemma3}        & 14.8 & 17.4 & \textbf{67.8} & 21.8 & 15.1 & \textbf{63.1} & 19.5 & 15.5 & \textbf{65.0} \\
            \llm{Ministral3}    & 12.5 & 15.5 & \textbf{72.0} & 21.6 & 19.5 & \textbf{58.9} & 22.0 & 12.5 & \textbf{65.5} \\
            
            \bottomrule
      \end{tabular}%
    }
    \caption{Win-Rate comparison of LLMs in the best-of-3 selection task. \textit{Note:} BL: Baseline Prompt without personality, MBTI: MBTI Prompt, BFI: BFI Prompt.}
    \label{tab:winrate}
      
\end{table}

\paragraph{Personality Comparison.}
Tab.~\ref{tab:winrate} reveals that the choice of personality directly impacts rationale quality. The BFI strategy demonstrates better alignment with human cognitive patterns (55.4-70.4\%), ahead of both MBTI and Baseline groups, particularly in consistency. For instance, \llm{GPT-5.1} achieved 68.9-78.8\% win rate with the BFI prompt across metrics, while the MBTI only 13.5-16.5\%. In contrast, MBTI prompts show stronger consistency but weaker plausibility and specificity compared to the baseline. This suggests that the trait-based detail of BFI provides a more robust choice for simulating nuanced appraisals compared to the binary nature of MBTI~\cite{FURNHAM1996303}.

\paragraph{Model Scale.}
Cognitive soundness exhibits a clear dependency on model capacity. \llm{GPT-5.1} consistently achieved the highest win rates on BFI prompt, indicating that complex rationale generation requires large-scale models. Among open-source models, \llm{Gemma-3-12B}  shows greater stability than smaller LLMs, maintaining a >60\% preference in plausibility under BFI settings. This suggests that robust rationale generation is a capacity-intensive task for current LLMs. A detailed analysis of the data is provided in Appendix~\ref{appendix:rq3_data_analysis}.

\section{Conclusion}
\label{sec:Conclusion}

We introduced \textbf{Persona-E\textsuperscript{2}}, a human-grounded dataset for personality-conditioned emotion analysis. Reliability experiments demonstrate that PAG reflect personality effects. Analysis of affective appraisals reveals domain-specific patterns, such as social media acts as an ``Emotional Black Hole.''  We also identified distinct, personality-shaped processes in emotion elicitation. While LLMs can map semantic neighborhoods in predicting emotional shifts, they currently lack precision. Our comparisons underscore the importance of cyber‑social emotional tasks, where BFI outperformed MBTI. Finally, cognitive prompt improves personalized reasoning, especially in large‑scale models.

\section*{Limitations}
% Events Number & source(other than 3) limited: History, Classics,
% Translation bias
% limited annotator (different educated level, culture background, People Number)
% limited emotion label space
% （emotoin subjectivity）：hard reproduction of annotator-level emotion selection
We introduce Persona-E\textsuperscript{2}, a large-scale dataset explicitly grounded in personality traits to model reader-based emotional variation. However, the dataset has several limitations. We acknowledge that the diversity of event sources is constrained, and the current version is limited to Chinese and English texts, which may not fully capture emotional expressions in other domains, cultures, or languages. Likewise, the limited number of annotators should be taken into consideration when assessing population effectiveness. Furthermore, cross-lingual translation introduces additional challenges: differences in phrasal connotations can shift the perceived emotional meaning. Moreover, emotion perception is inherently subjective; thus, individual differences among annotators hinder the consistent reproduction of emotion labels. Moreover, there is an inherent gap between task-oriented annotators and spontaneous real-world users. The emotion label space adopts Ekman’s six basic emotions plus neutral as a strategic trade-off, though this categorical scheme is coarser than dimensional or open-vocabulary alternatives. Additionally, the “General Writer” labels rely on an external classifier~\cite{hartmann2022emotionenglish} due to the unavailability of original author ratings, introducing potential algorithmic bias that should be considered when interpreting writer-reader emotion gaps.

\section*{Ethics Considerations}

\paragraph{Subjectivity and Diversity in Annotation}
We explicitly acknowledge that emotional appraisal is inherently subjective and culturally situated. Unlike traditional paradigms that seek a single ground truth, our data collection protocol respects diverse interpretations. We required annotators to report their genuine emotional reactions based on their personality profiles, ensuring the dataset captures the variance of human experience rather than enforcing a potentially biased consensus.

\paragraph{Annotator Welfare and Consent}
We recruited 36 annotators through university channels. All annotations were conducted via a self-developed and user-friendly online crowdsourcing platform. Participants were fully informed of the research purpose and potential risks, the public nature of the resulting dataset, and their right to withdraw at any time. We strictly adhered to fair labor practices; compensation was calculated based on task duration and complexity, ensuring that it significantly exceeded the local minimum wage.

\paragraph{Privacy and Data Anonymization}
The textual data used in Persona-E\textsuperscript{2} originates from publicly available news, social media, and personal narratives. To protect the privacy of the original content creators, we implemented a rigorous anonymization pipeline. All Personally Identifiable Information (PII), including real names, specific locations, and user handles, was excluded from the research.

\paragraph{Institutional Review}
The experimental protocol, including data collection and annotator interaction, was reviewed and approved by our Institutional Review Board prior to the study's commencement.

\paragraph{Representation}
We acknowledge distinct limitations in coverage. Although our dataset contains approximately 3,000 diverse events, it does not fully represent all events. Our pool of 36 annotators, while providing high annotation density, represents a specific demographic (university-educated, aged 18–25) that may not perfectly reflect the global population, potentially introducing demographic biases. We also utilized LLMs for initial data filtering, acknowledging that this automation may introduce minor biases despite human oversight.

\paragraph{Responsible Use and Sensitivity}
This dataset, to be released under a license compatible with its research-only creation purpose, must be used solely for non-commercial research. Derivatives must not be deployed in real-world applications beyond research prototypes, especially in commercial contexts. This dataset contains narratives that may be sensitive to specific cultural, religious, or social contexts. We urge researchers to exercise caution when deploying models trained on this data, particularly in high-stakes applications such as mental health support or behavioral analysis. Users must be aware that the models may reproduce the specific appraisal patterns of our annotator pool, which should not be interpreted as universal cultural truths.

\paragraph{Open Access and Reproducibility}
To foster transparency and encourage further research in personality-aware NLP, we will release the Persona-E\textsuperscript{2} dataset under a license that permits research use while prohibiting malicious applications. We believe open access is essential for the community to scrutinize, validate, and build upon our findings responsibly.

\paragraph{Use of AI Assistants}
We utilized AI assistants (e.g., \llm{ChatGPT-4o}) to refine the clarity and grammar of the manuscript. All scientific claims, experimental designs, and data analyses were conducted and verified by the human authors.

\section*{Societal Impact}

\paragraph{Advancing Affective Computing}
Our work promotes a paradigm shift from writer-centric sentiment analysis to reader-based emotional appraisal. By introducing the Persona-E\textsuperscript{2} dataset, we encourage the research community to move beyond static, single-label classification and explore how personality traits shape diverse interpretations. This transition is essential for building more inclusive AI systems that respect individual differences in perception.

\paragraph{Bridging Cognition and AI}
By integrating cognitive theories with LLMs, we highlight the value of psychological grounding in NLP. Our findings demonstrate that structured personality constraints (specifically BFI) enhance model reasoning, opening new avenues for personalized human-computer interaction. This theoretical alignment offers a foundation for developing more empathetic agents in mental health support and personalized education.

\paragraph{Decoupling Cyber-Social Dynamics}
Our analysis reveals that emotional elicitation varies significantly across news, social media, and life narratives. This distinction suggests that future research must decouple virtual interactions from physical-world events. Understanding distinct mechanisms like the negativity bias in social media provides actionable insights for monitoring digital sentiment and mitigating online polarization.

\paragraph{Potential Risks}
While personalized appraisal enhances user experience, it carries risks. The ability to tailor content to specific personality profiles could be misused for targeted manipulation or to reinforce echo chambers. Furthermore, without careful constraints, models might over-generalize personality traits, leading to unintended stereotyping. Researchers must prioritize safety and fairness when deploying these persona-aware systems.

\section*{Acknowledgments}
This work was supported by the Guangdong Provincial Key Laboratory of Human Digital Twin under Grant 2022B1212010004. We would like to express our sincere gratitude to all the annotators for their diligent and meticulous efforts, whose commitment was essential to the construction of our datasets. We also thank members of the HAI Lab at SCUT, our external collaborators, and the anonymous reviewers for their insightful comments and suggestions.

\bibliography{custom}

\clearpage
\appendix

\section{Broad Related Work}
\label{appendix:Broad_Related_Work}
Research on event-elicited emotion analysis and personality-conditioned modeling lays the foundation for our work. We review these two areas below to contextualize our contributions.

\subsection{Event-Elicited Emotion Analysis}

Early work established the baseline for event-driven affective analysis. Classic psychometric efforts like ISEAR~\cite{scherer1994evidence} and its follow-ups~\cite{troiano2019crowdsourcing, troiano2022x, troiano2023dimensional} collected first-person narratives of life events, treating the event as the primitive stimulus for actual affective responses, rather than inferring emotion from lexical cues alone. Subsequent datasets expanded this scope to interpersonal and social commonsense scenarios~\cite{rashkin2018event2mind, sap2019socialiqa, forbes2020social}.
% 评价体系插入事件-评价-情绪
Subsequent research enriched this view by inserting a mediating cognitive layer: appraisal~\cite{hofmann2020appraisal}. Work inspired by appraisal theory (e.g., x-enVENT~\cite{troiano2022x}, crowd-enVENT~\cite{troiano2023dimensional}) refined this view by integrating appraisal variables into the analysis of the relationship between events and emotions. In short, the community introduced multi-dimensional appraisal annotations and demonstrated that modeling appraisal variables substantially improves the model interpretability and predictive power.

% 对writer和reader进行考虑
A parallel line of research shifts the emphasis from the author to the audience. Datasets such as EmoBank~\cite{buechel2017emobank}, GoodNewsEveryone~\cite{oberlander2020goodnewseveryone}, GERSTI~\cite{dang2021emotion} and subsequent news corpora began to highlight that readers construct responses based on context rather than inheriting the writer’s stance. Similarly, social media benchmarks~\cite{hu2024resemo, ding2024lcsep} approximate reader affect via comments and reactions.  Multimodal efforts (BU-NEmo~\cite{reardon2022bu}, eMotions~\cite{wu2025emotions}, iNews~\cite{hu2025inews}) further showed that visual context and reader profiles modulate affective responses. These works mark a crucial shift from \textbf{``what does the text express?''} to \textbf{``how does the text make people feel?''}

% 总结缺陷
Despite these advances, most reader-based resources rely on aggregated labels (e.g., majority vote) or weak signals (e.g., clicks, emojis), which obscure inter-individual variability. Furthermore, while they cover specific domains, few datasets span multiple event sources to study generalized emotional elicitation. As a result, it is necessary to build a wide-rangeing emotion elicitation dataset grounded in real human variability.

\subsection{Personality-shaped Affective Computing}
Research on personality–emotion interaction has evolved along three main axes: explicitly measured personality, implicitly inferred personality, and LLM-simulated persona profiles. Most personality traits are measured using MBTI~\cite{myers1962myers} or BFI~\cite{john2010handbook}. While the former is more prevalent in public discourse, the latter is more widely acknowledged in academic psychology.

\paragraph{Explicit and Implicit Personality.}
Datasets utilizing explicit self-report or psychometric questionnaires (e.g., PANDORA~\cite{jurkovic2021pandora}) provide reliable grounding for linking traits to emotional tendencies. By aligning social media activity with personality traits, multimodal and conversational resources such as PersonaTAB~\cite{inoue2025personatab} and EmotionLines~\cite{hsu2018emotionlines} have extended this paradigm to dialogues and audiovisual interactions. While valuable, these datasets primarily capture writer-side expressions rather than reader-side elicitation. Conversely, implicit methods bypass questionnaires and derive personality from textual expressions~\cite{gao2013improving, hu2024llm, shen2025emoperso, li2025eerpd}. While scalable, these approaches lack ground truth and often struggle to distinguish between stable traits and temporary states.

\paragraph{LLM-Simulated Personality.}
Recent studies explore simulating diverse reactions via persona-prompted LLMs~\cite{mao2024editing, tu2024charactereval, chen2024oscars, samuel2024personagym, li2025big5, bai2025scaling}. While richer prompts with profiles are introduced, the behavioral fidelity of simulated agents improves accordingly~\cite{bai2025scaling, hu2024quantifying}. However, emerging evidence suggests a ``personality illusion''~\cite{han2025personality}: models often mimic linguistic styles (e.g., sounding ``angry'') rather than adopting the underlying appraisal mechanisms. Crucially, the field lacks a dataset that grounds these simulations in real human data, preventing rigorous verification of whether models truly capture trait-driven emotional diversity and reason about the underlying psychological causes. Consequently, no existing resource offers a large-scale, human-grounded repository while systematically capturing cross-person variation in emotional responses with multi-stage quality control.

\label{sec:appendix}

\section{Collection Details}
\subsection{Collection Source}
\label{appendix:CollectionSource}
To construct a diverse and comprehensive dataset, we aggregated data from twelve distinct sources spanning formal news, social media discussions, and specific life experience narratives. Below we provide brief descriptions and access links for each source:

\begin{itemize}
    \item \textbf{ABC News}: A collection of English-language breaking news and headlines, providing concise event summaries and titles to represent Western media perspectives. \href{https://abcnews.go.com/abcnews}{[Link]}
    
    \item \textbf{BBC News}: A global news feed offering comprehensive coverage of international events, featuring headlines and brief abstracts useful for analyzing formal journalistic sentiment. \href{https://www.bbc.com/}{[Link]}
    
    \item \textbf{The Independent}: A source of independent British journalism, supplying diverse news articles that contribute to the variation in editorial stance and topic coverage. \href{https://www.independent.co.uk/news}{[Link]}
    
    \item \textbf{Today (Jinri Toutiao)}: An aggregate of trending news titles from China. This source reflects current domestic hot topics and utilizes popularity metrics to gauge public interest. We utilize a collection platform for convenience. \href{https://tophub.today/c/tech}{[Link]}
    
    \item \textbf{The Paper}: A reputable Chinese digital media outlet providing in-depth coverage of current affairs and historical events with a broad temporal span, enriching the dataset with long-tail topics. We utilize a collection platform for convenience. \href{https://tophub.today/c/tech}{[Link]}
    
    \item \textbf{Weibo}: Sourced from one of China's largest social media platforms, this subset includes both official news releases and real-time trending topics. We selected an official account to collect trending topics. \href{https://weibo.com/u/2656274875}{[Link]}
    
    \item \textbf{WeChat (Weixin)}: A vast collection of articles from WeChat Public Accounts. It covers a wide array of social topics and cultural nuances not always present in mainstream news. We utilize a collection platform for convenience. \href{https://tophub.today/c/tech}{[Link]}
    
    \item \textbf{Reddit}: A large-scale aggregation of user-posted discussions covering diverse life scenarios-spanning from interpersonal conflicts to moral dilemmas-from subreddits such as \textit{r/SocialAnxiety} and \textit{r/PetPeeves}. These posts naturally contain rich emotional undercurrents. \href{https://www.reddit.com/}{[Link]}
    
    \item \textbf{B.E. (Benign Existence)}: Sourced from \textit{r/BenignExistence}, this subset contains non-dramatic, mundane life records. It serves as a crucial baseline for identifying objective and neutral emotional states. \href{https://www.reddit.com/r/BenignExistence/}{[Link]}
    
    \item \textbf{FMylife}: A collection of short, first-person narratives describing unfortunate or awkward daily moments. It provides specific scenarios for modeling negative, embarrassed, or self-deprecating affective responses. \href{https://www.fmylife.com/}{[Link]}
    
    \item \textbf{IUTB (I Used To Believe)}: User submissions of childhood misconceptions and naive beliefs. This unique source captures scenarios evoking innocence, confusion, or nostalgia. \href{http://iusedtobelieve.com/}{[Link]}
    
    \item \textbf{KindLife}: Stories of authentic altruism collected from \textit{RandomActsOfKindness}. This subset supplements the dataset with positive emotional dimensions, such as gratitude, warmth, and admiration. \href{https://www.randomactsofkindness.org/kindness-stories}{[Link]}
\end{itemize}

\subsection{Data processing}
\label{appendix:Data_processing}
\subsubsection{NSFW Filter}
\label{appendix:nsfw_filter}
To ensure the safety and cleanliness of the dataset, we employed a strict dual-model filtration pipeline. Specifically, we utilized \model{Distilbert-NSFW}~\cite{albouzidi2023distilbertNSFW} and \model{Roberta-large-NSFW}~\cite{tostai2023nsfwLarge} to detect potential offensive content. A data sample was discarded if either model predicted it as "NSFW" (Not Safe For Work) with a confidence score exceeding the default threshold. This rigorous process minimizes the inclusion of explicit or harmful text.
\begin{itemize}
    \item \textbf{\model{Distilbert-NSFW}~\cite{albouzidi2023distilbertNSFW}}: eliasalbouzidi/distilbert-nsfw-text-classifier
    \item \textbf{\model{Roberta-large-NSFW}~\cite{tostai2023nsfwLarge}}: TostAI/nsfw-text-detection-large
\end{itemize}

\subsubsection{Multi-Dimensional LLM Scoring}
\label{appendix:Multi-Dimensional LLM Scoring}

To curate a dataset capable of eliciting diverse emotional responses, we employed \llm{Qwen3-Max}~\cite{qwen3max} to score candidate texts based on their psychological ``differential potential.'' The scoring criteria vary slightly across domains to reflect their unique characteristics.
\paragraph{News Domain.} We prioritized news content with significant societal impact and personal relevance. The scoring prompt focuses on the text's ability to trigger divergent reactions based on reader personality.

\begin{tcolorbox}[promptstyle,  title={News Domain}]
    
    \textbf{\# SYSTEM ROLE} \\
    Expert in Media Psychology and Personality. Analyzes news headlines as psychological stimuli to elicit trait-dependent emotional responses and cognitive appraisals.
    
    \smallskip % 使用标准间距命令
    \textbf{\# OBJECTIVE} \\
    Evaluate headlines as emotional triggers. Core Criterion: \textit{High Differential Potential}—the capacity to evoke divergent reactions across distinct personality profiles.
    
    \smallskip
    \textbf{\# EVALUATION DIMENSIONS (Scale: 1-5)}
    \begin{itemize} % 压缩列表间距
        \item \textbf{1. Emotional Arousal:} Intensity of the emotional provocation triggered by the headline.
        \item \textbf{2. Personality Variability:} Divergence in responses based on traits (e.g., Neuroticism, Optimism/Pessimism, or Risk Preference).
        \item \textbf{3. Emotional Implicitness:} Degree of subtlety in triggering emotions (implicit framing vs. explicit emotional labels).
        \item \textbf{4. Personal Relevance:} Perceived connection to the average reader's daily life and immediate concerns.
    \end{itemize}
    
    \smallskip
    \textbf{\# SCORING GUIDELINES} \\
    \textbullet~ \textbf{Prioritize Trait-Variance:} Focus on divergence caused by sensitivity or orientation (e.g., Sensitive vs. Rational). \\
    \textbullet~ \textbf{Exclude External Biases:} Disregard political, ideological, or regional affiliations.
    
    \smallskip
    \textbf{\# INPUT TEXT:} \texttt{"\{text\}"} 
    
    \smallskip
    \textbf{\# OUTPUT FORMAT (JSON ONLY)} \\
    \texttt{\{"dim": \{"score": int, "reason": "str"\}\}} 
\end{tcolorbox}

\paragraph{Social Media Domain.} We focused on content reflecting digital social dynamics. The scoring mechanism rewards texts that allow for multi-vocal interpretations (e.g., vague-booking or complex social signaling).
\begin{tcolorbox}[promptstyle,  title={Social Media Domain}]
    
    \textbf{\# SYSTEM ROLE} \\
    Expert in Cyberpsychology and Personality. Analyzes social media content (e.g., status updates, comments) as projective stimuli to capture trait-bound emotional and behavioral variance.
    
    \smallskip % 使用标准间距命令
    \textbf{\# OBJECTIVE} \\
    Evaluate content as a psychological probe. Core Criterion: \textit{High Discriminant Validity}—the capacity to reveal personality differences through divergent interpretations and engagement styles.
    
    \smallskip
    \textbf{\# EVALUATION DIMENSIONS (Scale: 1-5)}
    \begin{itemize} % 压缩列表间距
        \item \textbf{1. Emotional Arousal:} Intensity of triggers (e.g., social exclusion, validation-seeking) and their susceptibility to personality modulation.
        \item \textbf{2. Personality Variability:} Contrast in reactions based on Big Five (e.g., Introversion vs. Extroversion) and Attachment Styles (Anxious vs. Avoidant).
        \item \textbf{3. Emotional Implicitness:} Reliance on subtext, irony, or "vague-booking" to provide a projective canvas for the reader.
        \item \textbf{4. Social Ecological Validity:} Alignment with common digital social dynamics (e.g., "seen but unreplied," social comparison, FOMO).
    \end{itemize}
    
    \smallskip
    \textbf{\# SCORING GUIDELINES} \\
    \textbullet~ \textbf{Reward Multivocality:} Prioritize texts that allow for multiple, trait-dependent interpretations (e.g., perceived as "bragging" vs. "inspiring"). \\
    \textbullet~ \textbf{Penalize Moral Universalism:} Avoid high scores for content that triggers a uniform moral response (e.g., consensus on extreme injustice).
    
    \smallskip
    \textbf{\# INPUT TEXT:} \texttt{"\{text\}"} 
    
    \smallskip
    \textbf{\# OUTPUT FORMAT (JSON ONLY)} \\
    \texttt{\{"dim": \{"score": int, "reason": "str"\}\}} % 保持一致的极简 JSON
\end{tcolorbox}

\paragraph{Life Experience Domain.} We emphasized scenarios highly relevant to ordinary daily life, enabling readers to project their own memories. The evaluation centers on ecological validity and emotional implicitness.

\begin{tcolorbox}[promptstyle, title={Life Experience Domain}]
    
    \textbf{\# SYSTEM ROLE} \\
    Expert in Personality and Social Psychology. Analyzes life narratives as projective stimuli to elicit differentiated responses based on traits and attachment styles.
    
    \smallskip % 使用标准间距命令而非 vspace
    \textbf{\# OBJECTIVE} \\
    Evaluate text as a psychological stimulus. Core Criterion: \textit{High Differential Potential} (divergent responses across profiles).
    
    \smallskip
    \textbf{\# EVALUATION DIMENSIONS (Scale: 1-5)}
    \begin{itemize} % 压缩列表间距
        \item \textbf{1. Emotional Arousal:} Intensity and trait-based variance.
        \item \textbf{2. Personality Variability:} Response divergence based on Big Five and Attachment Styles.
        \item \textbf{3. Emotional Implicitness:} Use of subtext and narrative gaps requiring projection.
        \item \textbf{4. Ecological Validity:} Relatability to common life stressors.
    \end{itemize}
    
    \smallskip
    \textbf{\# SCORING GUIDELINES} \\
    \textbullet~ \textbf{Prioritize Variance:} Reward "Rorschach-like" texts. \\
    \textbullet~ \textbf{Penalize Uniformity:} Avoid socially scripted responses.
    
    \smallskip
    \textbf{\# INPUT TEXT:} \texttt{"\{text\}"} 
    
    \smallskip
    \textbf{\# OUTPUT FORMAT (JSON ONLY)} \\
    \texttt{\{"dim": \{"score": int, "reason": "str"\}\}} 
\end{tcolorbox}

\subsubsection{Source-dependent Filtration Thresholds}
\label{appendix:source_thres}
Given the diverse nature of our data sources—ranging from formal news articles to informal social media discussions—the noise levels vary significantly. To address this, we applied source-specific quality filtration thresholds. As detailed in Tab.~\ref{tab:source_distribution}, we set distinct cutoff values for different domains to balance data quality and retention rates. These Thresholds are empirical for the better efficiency of filtering. For instance, social media sources like Reddit generally required a more lenient threshold (4.7) compared to formal news sources (3.5) to accommodate their colloquial nature while still filtering out low-quality inputs.

\begin{table}[!htbp]
    \centering
    \begin{tabular}{@{}llcc@{}}
        \toprule
        \textbf{Domain} & \textbf{Source} & \textbf{Cutoff} & \textbf{Count} \\
        \midrule
        \multirow{7}{*}{News} 
         & ABC\_news & 3.5 & 130 \\
         & BBC\_news & 3.5 & 248 \\
         & Independent & 3.5 & 36 \\
         & Today & 4.0 & 314 \\
         & The Paper & 4.0  & 273 \\ % has been checked & revised
         & Weibo & 4.0  & 478 \\
         & WeChat & 4.0  & 37 \\
        \midrule
        \multirow{2}{*}{Social Media} 
         & SocialChem & 4.0  & 416 \\
         & Reddit & 4.7  & 440 \\
        \midrule
        \multirow{4}{*}{Life Experience} 
         & B.E. & 4.0  & 140 \\
         & FMylife & 4.0  & 507 \\
         & IUTB & 4.0  & 26 \\
         & KindLife & 4.0  & 66 \\
        \bottomrule
    \end{tabular}
    \caption{Source Distribution \& Source-dependent Filtration Thresholds. \textit{Note:} Independent (Independent News), Today (Today's Headlines), SocialChem (SocialChemistry), B.E. (BenignExistence), KindLfe (RandomActsofKindness).}
    \label{tab:source_distribution}
\end{table}

\subsection{Sub-category labels}
\label{appendix:sub_cate}

To establish a unified taxonomy for downstream analysis, we defined a closed set of labels for each domain. While news categories were derived from existing mainstream media sections, subcategories for life experiences and social media were synthesized via LLM summarization. The specific prompts used for classification are detailed below.

In the news domain, we selected the most frequent categories to formulate the taxonomy, covering Economics, Technology, Sports, Entertainment, Society, Health, International, Environment, and Education.

\begin{tcolorbox}[promptstyle, title={News Classification Prompt}]
    \textbf{\# SYSTEM ROLE} \\
    Expert News Editor specialized in precise content categorization. Tasked with mapping news articles to a single, most relevant domain from a predefined taxonomy.

    \smallskip
    \textbf{\# TAXONOMY DEFINITIONS}
    \begin{itemize}
        \item \textbf{Economy:} Financial markets, corporate reports, macroeconomics, trade, and industry trends.
        \item \textbf{Technology:} Internet, Artificial Intelligence, electronics, and scientific breakthroughs.
        \item \textbf{Sports:} Tournaments, athlete updates, team news, Olympics, and World Cup.
        \item \textbf{Entertainment:} Movies, music, celebrities, variety shows, and cultural activities.
        \item \textbf{Social:} Livelihood, crime, accidents, human interest stories, and community updates.
        \item \textbf{Health:} Diseases, medical care, public health, wellness, and mental health.
        \item \textbf{International:} International relations, diplomatic events, and global affairs.
        \item \textbf{Environment:} Climate change, conservation, natural disasters, and energy issues.
        \item \textbf{Education:} Schooling, education reform, academic research, and admissions.
    \end{itemize}

    \smallskip
    \textbf{\# CLASSIFICATION RULES} \\
    - Assign the \textbf{single most relevant} category. \\
    - If multiple domains overlap, prioritize the primary focus of the reporting.

    \smallskip
    \textbf{\# INPUT NEWS CONTENT:} \texttt{"\{news\_content\}"}

    \smallskip
    \textbf{\# OUTPUT CONSTRAINT} \\
    Output \textbf{ONLY} the category name as a plain text string (e.g., \texttt{Economic}). \\
    \textbf{DO NOT} provide explanations, preambles, or additional formatting.
\end{tcolorbox}

As for the Social Media domain, events are categorized into Self-Recording, Emotional Expression, Informational Sharing, Social Discussion, and Humor Expression, based on the communicative intent.
\begin{tcolorbox}[promptstyle, title={Social Media Classification Prompt}]
    
    \textbf{\# SYSTEM ROLE} \\
    Expert Social Media Content Analyst. Categorize posts based on communicative intent and narrative structure into five predefined taxonomies.

    \smallskip
    \textbf{\# CLASSIFICATION HIERARCHY (Descending Priority)} \\
    1. \textbf{Humor Expression} $\rightarrow$ 2. \textbf{Social Discussion} $\rightarrow$ 3. \textbf{Informational Sharing} $\rightarrow$ 4. \textbf{Self-Recording vs. Emotional Expression}.

    \smallskip
    \textbf{\# TAXONOMY DEFINITIONS}
    \begin{itemize}
        \item \textbf{1. Self-Recording:} Descriptions of personal life events, trajectories, or interpersonal interactions. Focuses on \textit{narrative} (what happened).
        \item \textbf{2. Emotional Expression:} Pure subjective venting, state updates, or fragmented opinions. Focuses on \textit{internal states/stances} rather than event sequences.
        \item \textbf{3. Informational Sharing:} Fact-based content providing objective value (e.g., tutorials, guides, alerts). Characterized by high altruism.
        \item \textbf{4. Social Discussion:} Topics transcending the personal sphere (e.g., societal phenomena, public policy, collective ethics). From "I" to "Society."
        \item \textbf{5. Humor Expression:} Jokes, parodies, or ironic content intended primarily to entertain. Takes precedence if the core intent is comedic.
    \end{itemize}

    \smallskip
    \textbf{\# INPUT:} \texttt{Title: "\{title\}"; Content: "\{content\}"}

    \smallskip
    \textbf{\# OUTPUT FORMAT (JSON ONLY)} \\
    \texttt{\{} \\
    \indent \texttt{"category\_id": int,} \\
    \indent \texttt{"category\_name": "string",} \\
    \indent \texttt{"reasoning": "short\_explanation"} \\
    \texttt{\}}
\end{tcolorbox}

In the life experience domain, we defined five categories to classify events based on their behavioral nature: Interpersonal Interaction, Norm Transgression, Pursuit Consequences, Reputation Appraisal, and Routine Daily.
\begin{tcolorbox}[promptstyle, title={Life Experience Classification Prompt}]
    
    \textbf{\# SYSTEM ROLE} \\
    Expert Annotator for Behavioral Life Narratives. Tasked with mapping specific life experiences to a predefined five-class taxonomy based on their social-psychological core.

    \smallskip
    \textbf{\# TAXONOMY DEFINITIONS}
    \begin{itemize}
        \item \textbf{1. Interpersonal Interaction:} Focuses on the dynamics between $\geq 2$ parties (e.g., conflict, support, intimacy). Priority is given to the \textit{process} of interaction.
        \item \textbf{2. Norm Transgression:} Focuses on ethical or procedural violations (e.g., dishonesty, rule-breaking, moral dilemmas). Priority is given to \textit{transgression} over context.
        \item \textbf{3. Pursuit Consequences:} Focuses on the outcomes of goal-oriented tasks (e.g., success/failure in exams, career, or social attempts). Priority is given to \textit{achievement valence}.
        \item \textbf{4. Reputation Appraisal:} Focuses on social labeling and moral standing (e.g., gossip, being judged as "selfish" or "helpful"). Priority is given to \textit{public image}.
        \item \textbf{5. Routine Daily:} Focuses on mundane, low-tension activities (e.g., commuting, dining) without significant conflict or moral stakes.
    \end{itemize}

    \smallskip
    \textbf{\# CLASSIFICATION GUIDELINES} \\
    - Categorize based on the \textbf{primary narrative axis}. \\
    - For interpersonal transgressions, prioritize \textit{Norm Transgression}. \\
    - For goal-related embarrassments, prioritize \textit{Pursuit Consequences}.

    \smallskip
    \textbf{\# INPUT NARRATIVE:} \texttt{"\{text\}"}

    \smallskip
    \textbf{\# OUTPUT CONSTRAINT} \\
    Output \textbf{ONLY} the exact category name string from the list above (e.g., \texttt{Interpersonal Interaction}). \\
    \textbf{DO NOT} include numbers, explanations, JSON, or punctuation.
\end{tcolorbox}

\subsection{Dataset Examples}
\label{appendix:Dataset_Examples}
\begin{tcolorbox}[promptstyle, breakable, title={English Examples from Persona-E\textsuperscript{2}}]
\footnotesize 
1. Dr Blaine McGraw is alleged to have secretly filmed intimate videos of patients in his care. \\
2. The teacher, Abby Zwerner, was shot in January 2023 in her classroom at Richneck Elementary School in Newport News, Virginia. \\
3. Aircraft ‘disappeared from radar without transmitting distress signal’ minutes after entering Georgian airspace. \\
4. A woman sworn in as a city council member in Bangor, Maine, served time in prison for manslaughter. \\
5. A fire broke out at the venue hosting U.N. climate talks in Brazil, prompting evacuations as firefighters rushed to control the flames. \\
6. Today, we got back from our second honeymoon and went to pick the kids up from my mom's. Surprisingly, they were both sat quietly watching TV. Half jokingly, I asked my mom what her secret was. Without even a guilty pause she told me, Benadryl for chesty coughs in their juice. You're welcome. \\
7. Today, I went to the store for some pads with my dad. We got them and then went to the cashier. That's when he realized that they were scented. He took one out of the box, sniffed it, made me sniff it, then insisted the cashier smell it. \\
8. Today, I told my boyfriend I wouldn't be able to get any time off work to go to Mexico with him, and that we'd have to get our tickets refunded, and reschedule. He said not to bother, and that he already had someone else in mind to take with him. \\
9. Today, my best friend on Snapchat is my mum. \\
10. My mom recently passed away and I miss her more than words could ever express. During the height of the pandemic she underwent awful, brutal rounds of chemo and never ever complained. I have a photo of her on my desk showing her true RandomActsofKindness, a day in the life of my mom. She's walking into treatment wearing a mask, her cute bald head in a cute cap, with a big bag of sweet treats for the chemo nurses - to let them know how much she appreciated them! \\
11. Why do I always feel like I'm going to die and run through disaster scenarios every time I speak publicly at work? \\
12. Has anyone else felt or seen “ghosts”? When other people don’t notice? I’ve walked with friends and seen someone keeping pace, in my peripherally, on the sidewalk across the street. When I look over, no one is there. I’ve dreamt of people who passed away hours before or after they do. They never know they are dead, so I end up having to tell them. Looking into their eyes, taking in their scent for a moment more and letting them know why they feel so confused. \\
13. My coworker complained about being broke then showed up with a designer bag the next day\
14. Bought the apartment across the street for my parents—yet a bowl of soup’s distance has turned into yesterday’s leftovers. \\
15. Is it normal to not give your roommate a heads up about a SO sleeping over for 5 day per week ? \\
\end{tcolorbox}
\section{Annotation Details}
\subsection{Annotation Platform}
\label{appendix:Annotation_Platform}
\begin{figure*}[t]
    \centering
    \includegraphics[width=\textwidth]{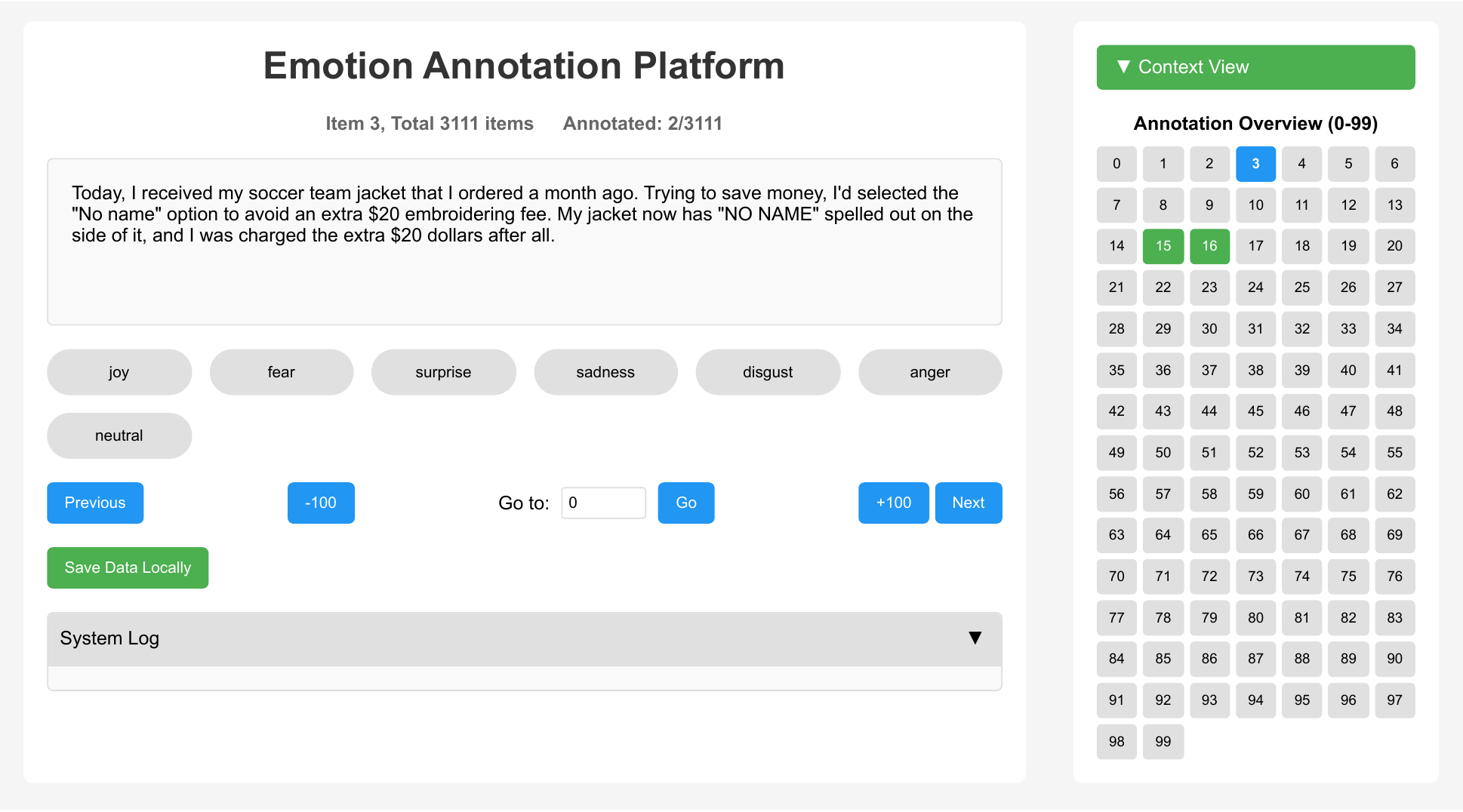}
    
    \caption{The annotation platform enables emotion labeling with seven categories, supporting live visualization, real-time data transmission, and comprehensive annotation monitoring.}
    
    \label{fig:Annotation Platform}
\end{figure*}

The annotation platform is designed for convenient online annotation and will be released in two months. The demonstration is shown in Fig.~\ref{fig:Annotation Platform}.

\subsection{Quality Control}
\label{sec:Quality_Control}
To ensure high-fidelity emotional annotations, we implemented a multi-layer quality control pipeline spanning annotator preparation, in-task monitoring, and post-hoc validation. All annotators underwent mandatory training before entering the task. The training clarified the central principle of this annotation scheme: annotators are required to report their stimulated emotion after reading the text, rather than infer the author’s sentiment or the event’s semantic polarity. When they realize their emotional reaction may diverge from what they perceive as the average response, annotators are explicitly instructed to record their genuine feeling, as inter-individual variability is essential to the study design. More specifically, each item is labeled with one primary emotion selected from 6 basic emotions plus a neutral class. Furthermore, the original English text is displayed alongside the translated Chinese version, allowing annotators to cross-reference and mitigate potential translation artifacts. All annotators possess advanced English reading proficiency.

During annotation, a real-time monitoring system captures behavioral traces—including latency, hesitation patterns, and abnormal repetition—which supports continuous quality auditing and early correction of potential low-engagement behaviors. We also tracked per-annotator throughput statistics, flagged abnormal labeling patterns (such as repeated use of the same emotion label or unrealistically rapid completion), and monitored longitudinal fatigue trends. Annotators exhibiting notable deviations received explicit reminders, and labels associated with confirmed anomalous behavior were re-annotated. This combination of structured training, reasoning-aligned instructions, and live supervision forms a robust quality assurance protocol and ensures that the final dataset reflects reliable, fine-grained reader-elicited emotional responses.

\subsection{Annotator Profiles}
\label{sec:Annotator_Profiles}
As shown in Tab.~\ref{tab:participant_profiles}, all annotators recruited are from China, and possess advanced proficiency in English reading comprehension and written expression. All 36 annotators (aged 18–25) are anonymized and indexed. Their personalities were profiled via MBTI and BFI questionnaires. MBTI denotes Myers--Briggs Type Indicator, where the four-letter codes represent Extraversion (E) / Introversion (I), Sensing (S) / Intuition (N), Thinking (T) / Feeling (F), and Judging (J) / Perceiving (P)~\cite{myers1962myers}, obtained from the MBTI-93 Questionnaire. BFI scores correspond to the Big Five personality dimensions (Openness, Conscientiousness, Extraversion, Agreeableness, Neuroticism)~\cite{john1991big}, obtained from the IPIP-NEO-120 questionnaire~\cite{JOHNSON201478} and normalized to $[0,1]$ based on the theoretical minimum and maximum scores of each dimension.
\begin{table}[t]
    \scriptsize
    \centering
    \begin{tabular}{@{}l@{}cccccccc}
        \toprule
        ID & GNDR & MBTI & Open. & Cons. & Extra. & Agree. & Neuro. \\
        \midrule
            E1  & M & INTP & 0.698 & 0.646 & 0.604 & 0.740 & 0.417 \\
            E2  & M & INFP & 0.615 & 0.354 & 0.479 & 0.552 & 0.677 \\
            E3  & F & INFP & 0.760 & 0.458 & 0.427 & 0.688 & 0.635 \\
            E4  & M & ISTJ & 0.500 & 0.802 & 0.500 & 0.729 & 0.406 \\
            E5  & M & INTJ & 0.688 & 0.792 & 0.510 & 0.781 & 0.312 \\
            E6  & F & ISTJ & 0.573 & 0.625 & 0.562 & 0.750 & 0.427 \\
            E7  & M & ENTP & 0.771 & 0.844 & 0.667 & 0.573 & 0.469 \\
            E8  & F & ISTJ & 0.656 & 0.750 & 0.490 & 0.635 & 0.448 \\
            E9  & M & ESFJ & 0.635 & 0.646 & 0.667 & 0.656 & 0.438 \\
            E10 & F & INFJ & 0.781 & 0.573 & 0.396 & 0.854 & 0.844 \\
            E11 & F & ESFJ & 0.760 & 0.781 & 0.646 & 0.802 & 0.188 \\
            E12 & F & ESTJ & 0.344 & 0.531 & 0.646 & 0.604 & 0.594 \\
            E13 & M & ESTP & 0.562 & 0.625 & 0.625 & 0.573 & 0.448 \\
            E14 & F & INTJ & 0.865 & 0.906 & 0.583 & 0.604 & 0.396 \\
            E15 & F & ENFP & 0.698 & 0.562 & 0.583 & 0.875 & 0.552 \\
            E16 & F & ENFP & 0.677 & 0.844 & 0.656 & 0.833 & 0.198 \\
            E17 & M & ISTJ & 0.625 & 0.552 & 0.427 & 0.510 & 0.844 \\
            E18 & F & ENTJ & 0.760 & 0.906 & 0.667 & 0.792 & 0.177 \\
            E19 & M & ESTJ & 0.500 & 0.625 & 0.646 & 0.771 & 0.406 \\
            E20 & M & ISTJ & 0.500 & 0.562 & 0.438 & 0.677 & 0.448 \\
            E21 & M & ISTJ & 0.729 & 0.896 & 0.625 & 0.802 & 0.188 \\
            E22 & F & ESTJ & 0.573 & 0.875 & 0.542 & 0.792 & 0.229 \\
            E23 & M & ESTP & 0.479 & 0.656 & 0.552 & 0.594 & 0.542 \\
            E24 & F & ISTJ & 0.552 & 0.771 & 0.562 & 0.667 & 0.240 \\
            E25 & M & INTJ & 0.781 & 0.781 & 0.562 & 0.719 & 0.146 \\
            E26 & M & ENTJ & 0.562 & 0.833 & 0.656 & 0.615 & 0.312 \\
            E27 & M & ENTJ & 0.833 & 0.854 & 0.823 & 0.688 & 0.208 \\
            E28 & M & ESTP & 0.646 & 0.677 & 0.604 & 0.781 & 0.177 \\
            E29 & M & ESTJ & 0.771 & 0.812 & 0.615 & 0.885 & 0.062 \\
            E30 & M & ISFP & 0.615 & 0.646 & 0.417 & 0.542 & 0.646 \\
            E31 & M & ISTJ & 0.500 & 0.646 & 0.417 & 0.604 & 0.396 \\
            E32 & F & INTP & 0.792 & 0.740 & 0.438 & 0.677 & 0.271 \\
            E33 & M & ISTJ & 0.677 & 0.854 & 0.562 & 0.812 & 0.385 \\
            E34 & F & ENTJ & 0.583 & 0.885 & 0.656 & 0.812 & 0.208 \\
            E35 & F & INTP & 0.635 & 0.552 & 0.521 & 0.771 & 0.521 \\
            E36 & M & ESTJ & 0.562 & 0.750 & 0.479 & 0.656 & 0.312 \\
        \bottomrule
    \end{tabular}
    \caption{Demographic and Personality Profiles of Participants. Annotators are indexed. BFI scores correspond to \textit{Openness, Conscientiousness, Extraversion, Agreeableness, Neuroticism}.}
    \label{tab:participant_profiles}
\end{table}

\section{Implementation Details}
\label{appendix:Implementation_Details}
On the cloud computing platform, the experiments for \textbf{RQ2} and \textbf{RQ3} required approximately 6 and 40 GPU hours on NVIDIA A100 GPUs, respectively.
\subsection{Hyper-parameters}
For both open-source and closed-source LLMs, we adjusted the generation hyper-parameters. For \textbf{RQ2}, we set temperature=0.2 to ensure stable emotion prediction. For \textbf{RQ3}, we set temperature=0.7 to facilitate reasoning about nuanced emotional reactions. Other hyper-parameters were kept at default values (max\_new\_tokens=1024, top\_p=0.9).
\subsection{Open-Source LLMs Versions}
The specific versions and Hugging Face identifiers for the open-source LLMs used in this study are listed below.
\begin{itemize}
    \item \textbf{\model{Meta-Llama-3-8B}~\cite{dubey2024llama}}: meta-llama/Meta-Llama-3-8B-Instruct
    \item \textbf{\model{Qwen3-8B}~\cite{yang2025qwen3}}: Qwen/Qwen3-8B
    \item \textbf{\model{Gemma-3-12B}~\cite{team2025gemma}}: google/gemma-3-12b-it
    \item \textbf{\model{Ministral-3-8B}~\cite{mistral2025ministral3hf}}: mistralai/Ministral-3-8B-Instruct-2512
\end{itemize}

\section{Experiment Details}
\subsection{Stability Analysis of Personality Clustering}
\label{appendix:Clustering_Stability}

To verify that the observed Personality Agreement Gap (PAG) is not sensitive to the specific choice of the clustering algorithm or the number of clusters ($k$), we evaluated three different methods: K-means, Gaussian Mixture Models (GMM), and Hierarchical Clustering. We varied $k$ from 3 to 9.

As shown in Table \ref{tab:pag_sensitivity}, the average PAG remains positive and significant across all configurations.

\begin{table}[th!]
\centering
\resizebox{\columnwidth}{!}{%
\begin{tabular}{lccccccc}
\toprule
\textbf{Method} & \textbf{k=3} & \textbf{k=4} & \textbf{k=5} & \textbf{k=6} & \textbf{k=7} & \textbf{k=8} & \textbf{k=9} \\
\midrule
K-Means & 10.45 & 13.45 & 13.92 & 16.44 & 18.10 & 20.08 & 23.10 \\
GMM & 12.00 & 12.95 & 14.85 & 18.70 & 20.29 & 19.86 & 22.24 \\
Hierarchical & 10.07 & 13.87 & 15.26 & 17.35 & 17.42 & 18.57 & 19.72 \\
\bottomrule
\end{tabular}%
}
\caption{Sensitivity analysis of the Personality Agreement Gap (PAG) across different clustering methods and cluster counts ($k$). All values represent the average Top-1 agreement gain (\%) over the global baseline.}
\label{tab:pag_sensitivity}
\end{table}

\subsection{RQ1. Dataset Affective Divergence}
(1) \textbf{General Writer (GW)}:Semantic sentiment of General Writer is predicted by a pre-trained emotion classifier~\cite{hartmann2022emotionenglish} that provides an identical label space (Ekman’s six basic emotions plus neutral) to our human annotations, ensuring direct comparability. (2) \textbf{General Reader (GR)}: majority-vote elicitation from all the annotations; and (3) \textbf{Persona Reader (PR)}: trait-conditioned elicitation from each personality cluster.
\subsubsection{General Writer vs. General Reader}
\label{appendix:rq1.1}
For General Writer, the available writer-side classifiers or domain-adapted models are not sufficient for performance comparison, since the label space of most existing models is incompatible with the one utilized in this study.
\paragraph{Results: } We analyzed the affective transition matrices between the GW and GR~\cite{buechel2017readers, troiano2019crowdsourcing}. As demonstrated in Fig~\ref{fig:three_heatmaps_row}, the three domains exhibit distinct polarity patterns in affective contagion (resonance) and shift (transfer).

Secondly, social media exhibits a sharp negativity bias: 81.6\% of neutral and 59.35\% of positive GW sentiments transfer into negative GR reactions, far exceeding news (27.6\%, 25.6\%) and life (30.8\%, 32.1\%) (Tab.~\ref{tab:rq1.1_polar2}). In contrast, life experience narratives demonstrate a strong positive shift, where 43.3\% of negative and 56.2\% of neutral GW sentiments transfer to positive GR emotions, significantly higher than news (28.7\%, 21.1\%) and social media (4.7\%, 11.8\%) (Shown in Tab.~\ref{tab:rq1.1_polar3}).

% --- 1. 定义颜色 ---
\definecolor{LowColor}{HTML}{BCE4D8}  % 浅色 (低分)
\definecolor{HighColor}{HTML}{2C5985} % 深色 (高分)

% --- 2. 设定阈值 ---
\def\MinVal{6} 
\def\MaxVal{81} 

% --- 3. 宏定义 ---
\newcommand{\AutoCell}[1]{%
    % [步骤1] 计算整数比例 (0-100)
    \pgfmathparse{int(round(100*(#1-\MinVal)/(\MaxVal-\MinVal)))}%
    \let\Ratio\pgfmathresult
    % 边界限制
    \ifnum\Ratio<0 \def\Ratio{0} \fi
    \ifnum\Ratio>100 \def\Ratio{100} \fi
    %
    % [步骤2] 关键修复：使用 \edef 构造命令
    % 这一步会强制把 \Ratio 变成具体的数字 (比如 50)，
    % 生成一个临时的命令 \MyColorCmd，内容不仅是颜色，而是包含了 \cellcolor...
    \edef\MyColorCmd{\noexpand\cellcolor{HighColor!\Ratio!LowColor}}%
    %
    % [步骤3] 执行这个构造好的命令
    \MyColorCmd%
    %
    % [步骤4] 字体颜色逻辑
    \ifnum\Ratio>40 \color{white} \else \color{black} \fi
    %
    % [步骤5] 输出数值
    #1%
}
\paragraph{Insight A}
In the News domain, a significant proportion of writer-expressed emotions transfers to neutral, while polar emotions maintain a relatively balanced resonance rate and high neutrality transfer (Tab.~\ref{tab:rq1.1_polar1}). Compared with the other domains, resonance rate-48.10\%, 43.66\% and 56.60\%-shows much more stability. Moreover, neutrality transfer-22.33-43.66\%- is significantly highest among 3 domains.
\begin{table}[h]
    \centering
    % ! 表示高度自动保持比例
    \resizebox{\columnwidth}{!}{%
        \renewcommand{\arraystretch}{1.2} % 保持行高舒适度
        \begin{tabular}{l c c c}
            \arrayrulecolor{black!30}\hline
            \multicolumn{4}{c}{\cellcolor{gray!10}\textbf{News}} \\
            % 顶部灰色细线
            \arrayrulecolor{black!30}\hline 
            \arrayrulecolor{black} % 恢复黑色供文字使用
             % & \textbf{Prompt v1} & \textbf{Prompt v2} & \textbf{Prompt v3} \\ 
            % 头部下方灰色细线
            \arrayrulecolor{black!30}\hline
            
            \textbf{Writer $\backslash$ Reader} & \textbf{Positive} & \textbf{Neutral} & \textbf{Negative} \\ 
            \arrayrulecolor{black!30}\hline
            \textbf{Positive} & \AutoCell{48.10} & \AutoCell{26.30} & \AutoCell{25.60} \\ 
            \textbf{Neutral} & \AutoCell{28.72} & \AutoCell{43.66} & \AutoCell{27.62} \\ 
            \textbf{Negative} & \AutoCell{21.07} & \AutoCell{22.33} & \AutoCell{56.60} \\ 
            
            % 底部灰色细线
            \arrayrulecolor{black!30}\hline
        \end{tabular}%
    }
    \caption{Affective polarity transition matrices between General Writer and General Reader in news domain.}
    \label{tab:rq1.1_polar1}
\end{table}

\paragraph{Insight B}
Conversely, Social Media exhibits a sharp negativity bias: 81.60\% of neutral and 59.35\% of positive GW sentiments transfer into negative GR reactions, while the positive and neutral sentiment rate are largely less than 20\%. 
\begin{table}[h]
    \centering
    % ! 表示高度自动保持比例
    \resizebox{\columnwidth}{!}{%
        \renewcommand{\arraystretch}{1.2} % 保持行高舒适度
        \begin{tabular}{l c c c}
            \arrayrulecolor{black!30}\hline
            \multicolumn{4}{c}{\cellcolor{gray!10}\textbf{Social Media}} \\
            % 顶部灰色细线
            \arrayrulecolor{black!30}\hline 
            \arrayrulecolor{black} % 恢复黑色供文字使用
             % & \textbf{Prompt v1} & \textbf{Prompt v2} & \textbf{Prompt v3} \\ 
            % 头部下方灰色细线
            \arrayrulecolor{black!30}\hline
            
            \textbf{Writer $\backslash$ Reader} & \textbf{Positive} & \textbf{Neutral} & \textbf{Negative} \\ 
            \arrayrulecolor{black!30}\hline
            \textbf{Positive} & \AutoCell{22.25} & \AutoCell{18.40} & \AutoCell{59.35} \\ 
            \textbf{Neutral} & \AutoCell{4.70} & \AutoCell{13.70} & \AutoCell{81.60} \\ 
            \textbf{Negative} & \AutoCell{11.78} & \AutoCell{15.35} & \AutoCell{72.87} \\ 
            
            % 底部灰色细线
            \arrayrulecolor{black!30}\hline
        \end{tabular}%
    }
    \caption{Affective polarity transition matrices between General Writer and General Reader in social media domain.}
    \label{tab:rq1.1_polar2}
\end{table}

\paragraph{Insight C}
In contrast, life experience narratives demonstrate a strong positive shift, where 43.28\% of negative and 56.20\% of neutral GW tones transfer to positive GR emotions.
\begin{table}[h]
    \centering
    % ! 表示高度自动保持比例
    \resizebox{\columnwidth}{!}{%
        \renewcommand{\arraystretch}{1.2} % 保持行高舒适度
        \begin{tabular}{l c c c}
            \arrayrulecolor{black!30}\hline
            \multicolumn{4}{c}{\cellcolor{gray!10}\textbf{Life Experience}} \\
            % 顶部灰色细线
            \arrayrulecolor{black!30}\hline 
            \arrayrulecolor{black} % 恢复黑色供文字使用
             % & \textbf{Prompt v1} & \textbf{Prompt v2} & \textbf{Prompt v3} \\ 
            % 头部下方灰色细线
            \arrayrulecolor{black!30}\hline
            
            \textbf{Writer $\backslash$ Reader} & \textbf{Positive} & \textbf{Neutral} & \textbf{Negative} \\ 
            \arrayrulecolor{black!30}\hline
            \textbf{Positive} & \AutoCell{62.40} & \AutoCell{5.55} & \AutoCell{32.05} \\ 
            \textbf{Neutral} & \AutoCell{56.20} & \AutoCell{13.00} & \AutoCell{30.80} \\ 
            \textbf{Negative} & \AutoCell{43.28} & \AutoCell{6.65} & \AutoCell{50.07} \\ 
            
            % 底部灰色细线
            \arrayrulecolor{black!30}\hline
        \end{tabular}%
    }
    \caption{Affective polarity transition matrices between General Writer and General Reader in life experience domain.}
    \label{tab:rq1.1_polar3}
\end{table}

\subsubsection{General Reader vs. Persona Reader}
\label{sec:rq1_r2r}
Personality traits significantly modulate transfer patterns. We use the acronym O, C, E, A, N to represent Openness, Conscientiousness, Extraversion, Agreeableness, and Neuroticism.
\paragraph{Insight A}
\label{sec:rq1_r2r_insight_a}
In the social media domain, we observe "Anxious Empathy" effect in Cluster 1 (High-A, High-N): neutral-to-negative transfer rate reaches 51.56\% and positive-to-negative 37.4\%, outstripping Cluster 3 (High-A, Low-N, 34.38\%, 19.19\%) and Cluster 4 (Low-A, High-N, 28.1\%, 19.2\%). Concrete data is shown in Tab.\ref{tab:rq1_r2r_insight_a}.

\begin{table}[t]
    \centering
    \small % 或者 \footnotesize 根据空间调整
    \renewcommand{\arraystretch}{1.2}
    % \setlength{\tabcolsep}{3pt}
    
    % 调整宽度以适应单栏
    \resizebox{\linewidth}{!}{
    \begin{tabular}{l ccc}
        \toprule
        \textbf{Start $\to$ Target} & \textbf{Positive} & \textbf{Neutral} & \textbf{Negative} \\
        \midrule
        
        % --- Cluster 1 ---
        \multicolumn{4}{c}{\cellcolor{gray!20}\textbf{Cluster 1: High A, High N}} \\ 
        % \cellcolor{gray!20} 是为了加一点背景灰度区分，需要 colortbl 宏包，不需要可去掉
        \textbf{Positive} & 45.45\% & 17.17\% & 37.37\% \\
        \textbf{Neutral}  & 11.72\% & 36.72\% & 51.56\% \\
        \textbf{Negative} &  3.45\% &  5.71\% & 90.84\% \\
        
        \midrule
        % --- Cluster 3 ---
        \multicolumn{4}{c}{\cellcolor{gray!20}\textbf{Cluster 3: High A, Low N}} \\
        \textbf{Positive} & 66.67\% & 14.14\% & 19.19\% \\
        \textbf{Neutral}  &  6.25\% & 59.38\% & 34.38\% \\
        \textbf{Negative} &  1.50\% &  6.31\% & 92.19\% \\
        
        \midrule
        % --- Cluster 4 ---
        \multicolumn{4}{c}{\cellcolor{gray!20}\textbf{Cluster 4: Low A, High N}} \\
        \textbf{Positive} & 77.78\% &  3.03\% & 19.19\% \\
        \textbf{Neutral}  & 19.53\% & 52.34\% & 28.12\% \\
        \textbf{Negative} &  4.20\% &  6.46\% & 89.34\% \\
        
        \bottomrule
    \end{tabular}
    }
    \caption{Polarity Transfer Matrix for different clusters in the Social domain. \textit{Note:} Grouped by cluster types. A: Agreement, N: Neuroticism.}
    \label{tab:rq1_r2r_insight_a}
\end{table}

\paragraph{Insight B}
\label{sec:rq1_r2r_insight_b}
A Negative Passivation effect appears in Cluster 0 (High-C, High-E), which exhibits the lowest negative resonance (73.3-80.6\% across domains) compared to the negative locking seen in Cluster 2 (Low-C, 87.3-92.5\%) and Cluster 3 (Low-E, 87.3-92.5\%). Concrete data is shown in Tab.\ref{tab:rq1_r2r_insight_b}.

\begin{table}[t]
    \centering
    \small % 或者 \footnotesize 根据空间调整
    \renewcommand{\arraystretch}{1.2}
    % \setlength{\tabcolsep}{6pt}
    
    % 调整宽度以适应单栏
    \resizebox{\linewidth}{!}{
    \begin{tabular}{l ccc}
        \toprule
        \textbf{Start $\to$ Target} & \textbf{Positive} & \textbf{Neutral} & \textbf{Negative} \\
        \midrule
        
        % --- Cluster 0 ---
        \multicolumn{4}{c}{\cellcolor{gray!20}\textbf{Cluster 0: High C, High E}} \\ 
        Positive & 66.57\% & 15.88\% & 17.55\% \\
        Neutral  & 27.98\% & 53.28\% & 18.73\% \\
        Negative & 12.55\% & 14.10\% & 73.34\% \\
        
        \midrule
        % --- Cluster 2 ---
        \multicolumn{4}{c}{\cellcolor{gray!20}\textbf{Cluster 2: Lower C}} \\
        Positive & 82.73\% &  3.62\% & 13.65\% \\
        Neutral  & 25.06\% & 52.55\% & 22.38\% \\
        Negative &  4.65\% &  2.82\% & 92.52\% \\
        
        \midrule
        % --- Cluster 3 ---
        \multicolumn{4}{c}{\cellcolor{gray!20}\textbf{Cluster 3: Lower E}} \\
        Positive & 71.59\% & 17.83\% & 10.58\% \\
        Neutral  & 11.19\% & 79.08\% &  9.73\% \\
        Negative &  9.45\% &  8.46\% & 82.09\% \\
        
        \bottomrule
    \end{tabular}
    }
    \caption{Polarity Transfer Matrix for different clusters in the News domain. \textit{Note:} Grouped by cluster types. C: Conscientiousness, E: Extraversion.}
    \label{tab:rq1_r2r_insight_b}
\end{table}

\paragraph{Insight C}
\label{sec:rq1_r2r_insight_c}
Openness shows a positive correlation with the Neutralization Rate ($r=+0.86$, $p=0.027$), where high-O individuals Cluster 0 transfer polar emotions to neutral at a rate of 88.2\%, while low-O individuals Cluster 2 do so at only 23.92\%. Concrete data is shown in Tab.\ref{tab:rq1_r2r_insight_c}.

\begin{table}[h]
    \centering
    % --- 样式设置 ---
    \normalsize

    % 使用 resizebox 确保表格宽度适配页面（例如占页面宽度的 90%）
    \resizebox{1\linewidth}{!}{
        \begin{tabular}{lcccccc} 
            \toprule
            \textbf{Metric} & \textbf{C0} & \textbf{C1} & \textbf{C2} & \textbf{C3} & \textbf{C4} & \textbf{C5} \\
            \midrule
            
            % 如果需要某一行高亮（模仿您之前的风格），可以取消下面这行的注释
            Open. & 0.823 & 0.718 & 0.553 & 0.695 & 0.556 & 0.612 \\
    
            Neu. Rate & 88.21\% & 64.28\% & 23.92\% & 58.72\% & 33.19\% & 70.75\% \\
            \bottomrule
        \end{tabular}
    }
    
    \caption{Comparison of Openness scores and Neutralization Rates across different clusters.}
    \label{tab:rq1_r2r_insight_c}
\end{table}

\subsection{RQ2. LLM Emotion Simulation}
\label{appendix:RQ2}
Before the construction of the SDS, we also conduct an experiment of randomly selected 100 events on \llm{GPT5.1}. In doing so, the results are shown in Tab.~\ref{tab:rq2_general_results}. The experiment reveals a significantly higher performance in social media compared with Tab.~\ref{tab:hard_set_results}. This suggests that \llm{GPT5.1} may fail to predict the emotional shifts of the social media content, highlighting the critical need of more emphasis on the cyber-social gap.
\subsubsection{SDS construction}
\subsubsection{Definition and Construction}
The SDS is constructed to capture scenarios where emotional responses are unambiguous within specific personality groups but contradictory between them. We filter events based on a dual-criteria mechanism:
\begin{enumerate}
    \item \textbf{Intra-group Consistency:} We retain events where specific personality groups demonstrate high internal agreement ($S_{consensus}(G) > \alpha, \alpha = 0..3$), ensuring the emotional signal is not random noise. Sensitivity analysis for $\alpha$ has not been conducted.
    \item \textbf{Inter-group Divergence:} Among high-consensus events, we select those where the dominant emotional labels differ significantly across distinct personality profiles.
\end{enumerate}
Following this process, the final SDS comprises 413 events, distributed across domains as 257 from News, 69 from Social Media, and 87 from Life Experience.
\subsubsection{Prompt Settings}
We conducted experiments using three distinct prompt strategies to evaluate the capabilities of LLM emotion prediction: General Prompt, Persona Prompt, and Persona-CoT. The specific templates and instructions for each strategy are detailed below.
\begin{tcolorbox}[promptstyle, breakable, title={General Reader Emotion Prediction}]    
    \textbf{\# SYSTEM ROLE} \\
    You are a general observer representing an average human reaction. Your task is to read an event and report the most natural immediate emotional reaction. Do NOT assume any specific personality traits (like Neuroticism or Extraversion).

    \smallskip
    \textbf{\# INPUT EVENT} \\
    \texttt{"[EVENT\_DESCRIPTION]"}

    \smallskip
    \textbf{\# VALID LABELS}
    \begin{itemize}
        \item \textbf{Emotions:} Joy, Fear, Surprise, Sadness, Disgust, Anger, Neutral.
        \item \textbf{Intensity:} 1 (Very Weak) to 5 (Very Strong).
    \end{itemize}

    \smallskip
    \textbf{\# INSTRUCTION} \\
    Return a JSON object containing the TOP 2 most likely emotions, ranked by confidence.

    \smallskip
    \textbf{\# OUTPUT FORMAT} \\
    Return exactly 2 emotions in descending order of confidence:
    \smallskip
    \texttt{\{}\\
    \texttt{\hspace*{0.5em}"predictions": [}\\
    \texttt{\hspace*{1em}\{"Emotion": "Joy", "Intensity": 4\},}\\
    \texttt{\hspace*{1em}\{"Emotion": "Surprise", "Intensity": 3\}}\\
    \texttt{\hspace*{0.5em}]}\\
    \texttt{\}}
\end{tcolorbox}

\begin{tcolorbox}[promptstyle, breakable, title={Persona-based Emotion Prediction}]
    \textbf{\# SYSTEM ROLE} \\
    You are a participant in a psychology experiment simulating a specific human persona. Your task is to read an event description and report your immediate emotional reaction based on your persona.

    \smallskip
    \textbf{\# PERSONA PROFILE} \\
    \texttt{[PERSONA\_DESCRIPTION]}

    \smallskip
    \textbf{\# INPUT EVENT} \\
    \texttt{"[EVENT\_DESCRIPTION]"}

    \smallskip
    \textbf{\# VALID LABELS}
    \begin{itemize}
        \item \textbf{Positive Candidates:} [Joy, Surprise]
        \item \textbf{Negative Candidates:} [Sadness, Anger, Fear, Disgust, Surprise]
        \item \textbf{Neutral Candidate:} [Neutral]
        \item \textbf{Intensity:} 1 (Very Weak) to 5 (Very Strong).
    \end{itemize}

    \smallskip
    \textbf{\# INSTRUCTIONS} \\
    \textbf{Step 1 -- Polarity assessment:} Based on your personality traits and the event description, decide whether the overall emotion is POSITIVE, NEGATIVE, or NEUTRAL. \\
    \textbf{Step 2 -- Final selection:} From the chosen category, pick the top two emotions that best match both your persona and the event.

    \smallskip
    \textbf{\# OUTPUT FORMAT} \\
    Return exactly 2 emotions in descending order of confidence:
    \smallskip
    \texttt{\{}\\
    \texttt{\hspace*{0.5em}"predictions": [}\\
    \texttt{\hspace*{1em}\{"Emotion": "Joy", "Intensity": 4\},}\\
    \texttt{\hspace*{1em}\{"Emotion": "Surprise", "Intensity": 3\}}\\
    \texttt{\hspace*{0.5em}]}\\
    \texttt{\}}
\end{tcolorbox}

\begin{tcolorbox}[promptstyle, breakable, title={CoT Persona Emotion Prediction}]
    \textbf{\# SYSTEM ROLE} \\
    You are a participant in a psychology experiment simulating a specific human persona. Your task is to read an event description and report your immediate emotional reaction based on your persona.

    \smallskip
    \textbf{\# PERSONA PROFILE} \\
    \texttt{[PERSONA\_DESCRIPTION]}

    \smallskip
    \textbf{\# INPUT EVENT} \\
    \texttt{"[EVENT\_DESCRIPTION]"}

    \smallskip
    \textbf{\# VALID LABELS}
    \begin{itemize}
        \item \textbf{Positive Candidates:} [Joy, Surprise]
        \item \textbf{Negative Candidates:} [Sadness, Anger, Fear, Disgust, Surprise]
        \item \textbf{Neutral Candidate:} [Neutral]
        \item \textbf{Intensity:} 1 (Very Weak) to 5 (Very Strong).
    \end{itemize}

    \smallskip
    \textbf{\# INSTRUCTIONS} \\
    \textbf{Step 1 -- Polarity assessment:} Based on your personality traits and the event description, decide whether the overall emotion is POSITIVE, NEGATIVE, or NEUTRAL. \\
    \textbf{Step 2 -- Final selection:} From the category chosen, pick the top two emotions that best match both your personality and the event. Report: Polarity, Top1 Emotion, and Top2 Emotion.

    \smallskip
    \textbf{\# OUTPUT FORMAT} \\
    Return exactly 2 emotions in descending order of confidence:
    \smallskip
    \texttt{\{}\\
    \texttt{\hspace*{0.5em}"predictions": [}\\
    \texttt{\hspace*{1em}\{"Emotion": "Joy", "Intensity": 4\},}\\
    \texttt{\hspace*{1em}\{"Emotion": "Surprise", "Intensity": 3\}}\\
    \texttt{\hspace*{0.5em}]}\\
    \texttt{\}}
    
    Let's think step by step.
\end{tcolorbox}

\subsubsection{Data Analysis}
We also provide the data when events are randomly selected from all 3113 dataset

\begin{table*}[t!]
  \centering
  \footnotesize
  \setlength\tabcolsep{2.5pt} % 稍微增加一点间距，如果太宽可改回 1.5pt
  % \resizebox{\textwidth}{!}{% % 如果表格过宽，可以取消注释这一行
      \begin{tabular}{l ccc ccc ccc ccc} % 1个左对齐列(Metric) + 12个居中列(数据)
        \toprule
        \multirow{2}{*}{\textbf{Metric}} 
        & \multicolumn{3}{c}{\textbf{News}} 
        & \multicolumn{3}{c}{\textbf{Social Media}} 
        & \multicolumn{3}{c}{\textbf{Life Experience}} 
        & \multicolumn{3}{c}{\textbf{Overall}} \\ 
        \cmidrule(lr){2-4} \cmidrule(lr){5-7} \cmidrule(lr){8-10} \cmidrule(lr){11-13}
        
        % 第二行表头：注意第一列为空，因为那是 Metric 占据的位置
        & \textbf{General} & \textbf{Persona} & \textbf{CoT}
        & \textbf{General} & \textbf{Persona} & \textbf{CoT}
        & \textbf{General} & \textbf{Persona} & \textbf{CoT}
        & \textbf{General} & \textbf{Persona} & \textbf{CoT} \\
        \midrule
        
        % --- Top-1 Accuracy ---
        Top-1 Accuracy
        & 34.0 & 32.0 & 34.0
        & 37.1 & 37.1 & 34.3
        & 33.3 & 46.7 & 40.0
        & 35.0 & 36.0 & 35.0 \\
        
        \midrule
        
        % --- Top-2 Accuracy ---
        Top-2 Accuracy
        & 62.0 & 60.0 & 58.0
        & 48.6 & 45.7 & 45.7
        & 53.3 & 53.3 & 46.7
        & 56.0 & 54.0 & 52.0 \\
        
        \bottomrule
      \end{tabular}%
  % }
  \caption{Performance of \llm{GPT5.1} on the randomly selected set from the dataset.}
  \label{tab:rq2_general_results}
\end{table*}

\label{appendix:rq2_data_analysis}
As detailed in Tab.~\ref{tab:hard_set_results}, the experiments on the Hard Set reveal distinct trends across model capacities, prompt strategies, and domains discrepancy. First, while the average Top-1 accuracy hovers around 25.0\%, Top-2 accuracy surges to approximately 45.0\%. Secondly, introducing persona constraints yields marginal gains for \llm{GPT-5.1} (29.0-31.0\%) but significantly boosts \llm{Qwen3-8B} (20.0-28.0\%). Conversely, smaller models like \llm{LLAMA3-8B} suffer slight performance degradation under complex prompts. Finally, models consistently underperform in the Social Media domain (GPT-5.1: 18.2-27.3\% for Top-1), lagging far behind News and Life Experience.

\subsection{RQ3. Cognitive Soundness}
\label{appendix:rq3}
We assess the cognitive plausibility of the reasoning process with 3 different types of prompt in random order (Baseline/MBTI/BFI Prompt). For each test instance, reviewers performed a best-of-three forced-choice selection for three metrics: a) Persona Consistency, b) Reasoning Plausibility, and c) Emotion Specificity
\subsubsection{Baseline Prompt}
\begin{tcolorbox}[promptstyle, breakable, title={Baseline Reasoning Prompt}]
    \textbf{\# SYSTEM ROLE} \\
    You are a cognitive psychology expert. Please analyze the connection between the text and the reported emotion solely based on the text content. Strictly adhere to these rules: 1) Output the analysis directly without fillers like "Okay"; 2) Do not output your thinking process; 3) Strictly follow the required format.

    \smallskip
    \textbf{\# USER ROLE} \\
    Please analyze the underlying reasons for the reported emotion and intensity when reading the text below based on the content of the text itself:

    \smallskip
    \texttt{[Event Text]: \{event\_text\}} \\
    \texttt{[Reported Emotion]: \{emotion\}} \\
    \texttt{[Emotion Intensity]: \{intensity\} (1-5)}

    \smallskip
    Please output the analysis in the following format:
    \begin{itemize}
        \item \textbf{Event Summary:} (A one-sentence summary)
        \item \textbf{Reasoning Process:} (Brief analysis based on the text content)
        \item \textbf{Identified Cause:} (Likely motivation derived from the text)
    \end{itemize}
\end{tcolorbox}

\subsubsection{MBTI Prompt}

\begin{tcolorbox}[promptstyle, breakable, title={MBTI-driven Reasoning Prompt}]
    \textbf{\# SYSTEM ROLE} \\
    You are a psychology expert. Analyze the user's emotions based on MBTI personality theory. \textbf{[Current Subject]} \{real\_name\} (MBTI Type: \{mbti\}). \textbf{[Rules]} 1) Output the analysis directly without fillers like "Okay"; 2) Do not output your thinking process; 3) Strictly follow the required output format.

    \smallskip
    \textbf{\# USER ROLE} \\
    Please analyze the underlying reasons for \{real\_name\}'s reported emotion and intensity when reading the text below:

    \smallskip
    \texttt{[Event Text]: \{event\_text\}} \\
    \texttt{[Reported Emotion]: \{emotion\}} \\
    \texttt{[Emotion Intensity]: \{intensity\} (1-5)}

    \smallskip
    Please output the analysis in the following format:
    \begin{itemize}
        \item \textbf{Event Summary:} (A one-sentence summary)
        \item \textbf{Reasoning Process:} (Brief analysis combining \{mbti\} traits)
        \item \textbf{Identified Cause:} (Psychological motivation)
    \end{itemize}
\end{tcolorbox}

\subsubsection{BFI Prompt}

\begin{tcolorbox}[promptstyle, breakable, title={Personality-driven Reasoning Prompt}]
    \textbf{\# SYSTEM ROLE} \\
    You are a cognitive psychology expert. Analyze the user's emotional response based on the Big Five Personality Traits (OCEAN) theory. 
    
    \smallskip
    \textbf{[Score Interpretation]} Normalized values from 0.0 to 1.0 (0.0: Lowest; 1.0: Highest; 0.5: Medium). 
    \textbf{[Dimensions]} \textbf{1. Openness (O):} creativity, intelligence; \textbf{2. Conscientiousness (C):} self-discipline, dutifulness; \textbf{3. Extraversion (E):} sociability, optimism; \textbf{4. Agreeableness (A):} altruism, empathy; \textbf{5. Neuroticism (N):} emotional instability, anxiety.

    \smallskip
    \textbf{[Current Subject]} Name: \{real\_name\}; Personality Scores: \{bfi\_desc\}.
    
    \smallskip
    \textbf{[Rules]} 1) Output analysis directly without fillers; 2) Explain how specific O/C/E/A/N dimensions influenced the emotion; 3) Do not output thinking process; 4) Follow the required format.

    \smallskip
    \textbf{\# USER ROLE} \\
    Please analyze the underlying reasons for \{real\_name\}'s reported emotion and intensity when reading the text below:

    \smallskip
    \texttt{[Event]: \{event\_text\}} \\
    \texttt{[Emotion]: \{emotion\}} \\
    \texttt{[Intensity]: \{intensity\} (1-5)}

    \smallskip
    Please output the analysis in the following format:
    \begin{itemize}
        \item \textbf{Event Summary:} (One sentence summary)
        \item \textbf{Personality Analysis:} (Explicitly mention which dimension (O/C/E/A/N) scores played a key role.)
        \item \textbf{Final Attribution:} (Summary of psychological motivation)
    \end{itemize}
\end{tcolorbox}

After generating reasoning chains across different styles, we recruited five annotators to conduct a comparative evaluation using a "best-of-3" selection protocol. During the review process, each reviewer was required to perform a forced-choice selection based on three criteria: a) Persona Consistency, b) Reasoning Plausibility, and c) Emotional Specificity. The formal definitions of these metrics, used to evaluate the cognitive soundness of the LLM-generated reasoning, are provided below:

\begin{itemize}
    \item \textbf{Cognitive Consistency}: Assesses whether the rationale aligns with the specific values and behavioral patterns of the assigned persona. A rationale that exhibits thinking patterns divergent from those of the specified persona is considered of poor quality.
    \item \textbf{Reasoning Plausibility}: Evaluates the logical coherence of the causal chain from the event to the elicited emotion. A post-hoc justification that merely reverse-engineers the given emotion is considered poor quality.
    \item \textbf{Emotional Specificity}: Measures how precisely the rationale is tailored to explain the specific target emotion. A vague rationale that could explain any emotional response is considered poor quality.
\end{itemize}

\subsubsection{Data Analysis}
\label{appendix:rq3_data_analysis}
\paragraph{Personality Comparison.}
We conducted experiments on a) Baseline prompts without personality, b) MBTI prompt, c) BFI prompt. Also, we introduced the cognitive appraisal process into the personality prompts. As shown in Tab.~\ref{tab:winrate}, the BFI prompting strategy demonstrates a leading performance across all metrics, particularly in Persona Consistency (55.4-70.4\%), eclipsing the MBTI and Baseline groups. While MBTI marginally outperforms the baseline group in consistency, it unexpectedly underperforms in Reasoning Plausibility and Emotional Specificity.
\paragraph{Model Comparison.}
\llm{GPT-5.1} consistently achieves the highest win rates across all dimensions. Among open-source models, performance correlates with parameter scale: \llm{Gemma-3-12B} significantly outperforms smaller LLMs , maintaining >60\% preference in BFI settings. \llm{Qwen-8B} lags behind, particularly in specificity.

\end{document}